\documentclass{article}

\usepackage{PRIMEarxiv}

\usepackage{amsmath}
\usepackage[utf8]{inputenc} 
\usepackage[T1]{fontenc}    
\usepackage{hyperref}       
\usepackage{url}            
\usepackage{booktabs}       
\usepackage{amsfonts}       
\usepackage{nicefrac}       
\usepackage{microtype}      
\usepackage{lipsum}
\usepackage{fancyhdr}       
\usepackage{graphicx}       
\graphicspath{{media/}}     
\usepackage{todonotes}
\usepackage{amssymb}
\usepackage{makecell}

\title{WeirNet: A Large-Scale 3D CFD Benchmark for Geometric Surrogate Modeling of Piano Key Weirs}

\author{
  \textbf{Lisa Lüddecke\textsuperscript{1}, Michael Hohmann\textsuperscript{2, 3}, Sebastian Eilermann\textsuperscript{2, 3},} \\
  \textbf{Jan Tillmann-Mumm\textsuperscript{5}, Pezhman Pourabdollah\textsuperscript{4}, Mario Oertel\textsuperscript{1}, Oliver Niggemann\textsuperscript{2, 3}} \\
  \textsuperscript{1}Hydraulic Engineering Section, Civil Engineering, Helmut-Schmidt-University, Hamburg, Germany \\
    \textsuperscript{2}Institute for Artificial Intelligence, Helmut-Schmidt-University, Hamburg, Germany   \\
    \textsuperscript{3}Professorship of Computer Science in Mechanical Engineering, Helmut-Schmidt-University, Hamburg, Germany \\
   \textsuperscript{4}Chair for Automation Technology, Helmut-Schmidt-University, Hamburg, Germany \\
  \textsuperscript{5}Federal Waterways Engineering and Research Institute, Karlsruhe, Germany \\
  \textsuperscript{1,2,3,4}firstname.lastname@hsu-hh.de \\
  \textsuperscript{5}firstname.lastname@baw.de
  }

\begin{document}

\maketitle

\begin{abstract}
\textcolor{black}{Reliable prediction of hydraulic performance remains a central challenge in designing Piano Key Weirs because discharge capacity depends sensitively on three-dimensional geometry and operating conditions.} Data-driven surrogate models can accelerate hydraulic-structure design, but progress is limited by the scarcity of large, well-documented datasets that jointly capture geometric variation, operating conditions, and functional performance. This study presents \emph{WeirNet}, a large-scale 3D CFD benchmark dataset for geometric surrogate modeling of Piano Key Weirs. WeirNet contains 3,794 parametrically generated, feasibility-constrained rectangular and trapezoidal Piano Key Weir geometries, each scheduled at 19 discharge conditions using a consistent free-surface OpenFOAM\textregistered\ workflow, resulting in 71,387 successfully completed simulations that constitute the released benchmark and provide complete discharge coefficient labels. The dataset is released in multiple modalities compact parametric descriptors, watertight surface meshes and high-resolution point clouds together with standardized tasks and in-distribution and out-of-distribution splits. Representative surrogate families are benchmarked for discharge coefficient prediction. Tree-based regressors on parametric descriptors achieve the best overall accuracy, while point- and mesh-based models remain competitive and offer parameterization-agnostic inference. All surrogates evaluate in milliseconds per sample, providing orders-of-magnitude speedups over CFD runtimes. Out-of-distribution results identify geometry shift as the dominant failure mode compared to unseen discharge values, and data-efficiency experiments show diminishing returns beyond roughly 60\% of the training data. \textcolor{black}{By publicly releasing the dataset together with simulation setups and evaluation pipelines, WeirNet establishes a reproducible framework for data-driven hydraulic modeling and enables faster exploration of Piano Key Weir designs during the early stages of hydraulic planning.} 
\end{abstract}


\section{Introduction}
\label{sec:intro}

Dams and weirs are hydraulic control structures and critical safety components of river systems. To ensure safe operation under every possible discharge event, it is essential to reliably predict their hydraulic capacity and behavior across a wide range of operating conditions~\cite{Schleiss.2011}.

\begin{figure*}
    \centering
    \includegraphics[width=0.95\linewidth]{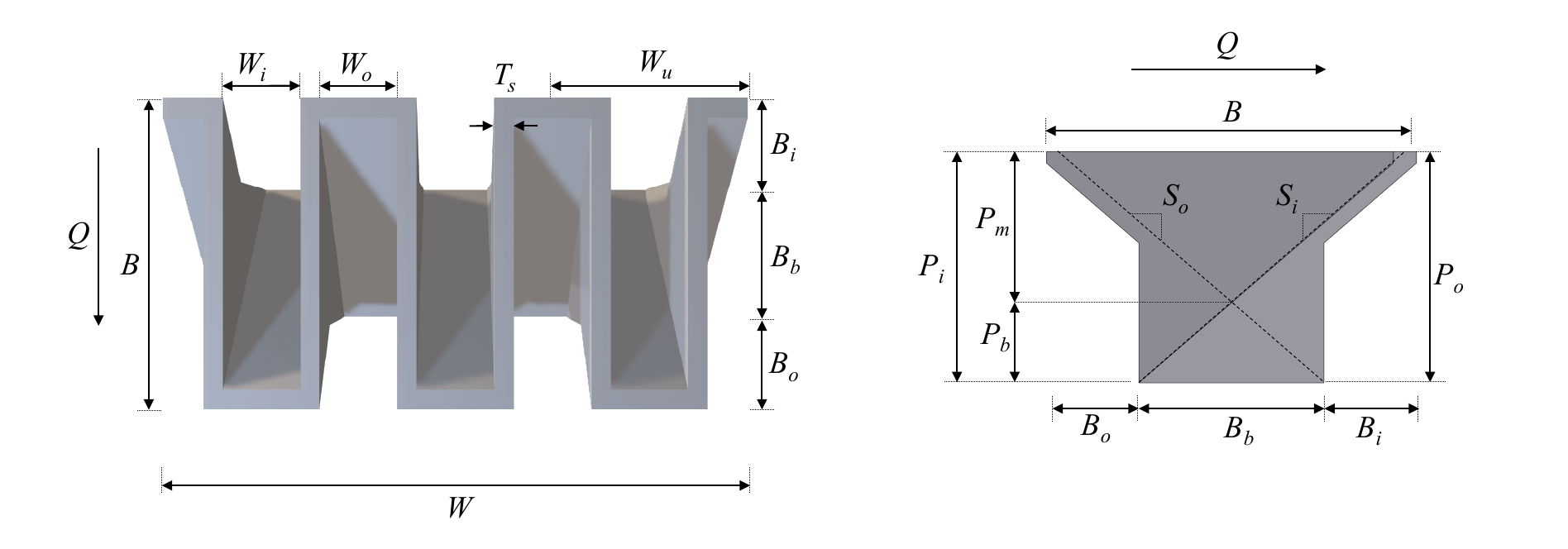}
    \caption{Main geometric parameters of a rectangular PKW in plan view (left) and side view (right) defined by Pralong et al.~\cite{pralong2011naming} with $Q$ indicating the direction of flow.}
    \label{fig:Pralong}
\end{figure*}

The Piano Key Weir (PKW) is a type of weir structure introduced in Lempérière and Ouamane~\cite{Lemperiere.2003} as a solution to increase discharge capacity compared to regular linear weirs by extending the centerline crest length. Since the discharge efficiency of a weir is directly proportional to the overfall length, the PKW achieves significantly higher discharge coefficients due to its folded design, which multiplies the crest length for the same clear width. Laboratory studies have shown that the discharge over a Piano Key Weir can be 2 to 3 times greater than that over a linear broad-crested weir with the same clear width $W$~\cite{LeiteRibeiro.2012,Schleiss.2011}. This weir differs from its predecessor, the Labyrinth weir, by having a reduced footprint, achieved through inclined inlet and outlet keys that create upstream and downstream overhangs. These overhangs, along with the folded design, introduce several geometric parameters that define the weir’s structure. To standardize these parameters, Pralong et al.~\cite{pralong2011naming} proposed a nomenclature describing the main features of the Piano Key Weir, which is now commonly used. These parameters are illustrated in Figure \ref{fig:Pralong}. A comprehensive delineation of these parameters can be ascertained from Table \ref{TAB_Pralong}.

\begin{table}[tb]
\caption{Nomenclature of the parameters of the PKW geometry by Pralong et al.~\cite{pralong2011naming}.}
\label{TAB_Pralong}
\centering
\begin{tabular}{l l}
\toprule
Parameters &	 Definition \\
\midrule
$B$ & Upstream-downstream length of the PKW (total weir length)\\
$B_o$ & Upstream (outlet key) overhang crest length\\
$B_i$ & Downstream (inlet key) overhang crest length\\
$B_b$ & Base length\\
$P_i$ & Height of the inlet entrance measured from the PKW crest (including possible parapet walls)\\
$P_o$ & Height of the outlet entrance measured from the PKW crest (including possible parapet walls)\\
$P_b$ & Height of the apron level at inlet key and outlet key intersection\\
$P_m$ & Difference between $P_i$ and $P_b$\\
$S_i$ & Slope of the inlet key apron (length over height) \\
$S_o$ & Slope of the outlet key apron (length over height)\\
$W$ & Total width of the PKW\\
$W_u$ & Width of a PKW unit\\
$W_i$ & Inlet key width (sidewall to sidewall)\\
$W_o$ & Outlet key width (sidewall to sidewall)\\
$T_{s}$ & Sidewall thickness\\
$T_i$ & Horizontal crest thickness at inlet key extremity\\
$T_o$ & Horizontal crest thickness at outlet key extremity\\
$L$ & Total developed centerline length along the overflowing crest axis\\
$L_u$ & Developed length of the PKW unit along the overflowing crest axis\\
$N_u$ & Number of PKW units\\
\bottomrule
\end{tabular}
\end{table}

The complex geometry of the PKW results in a highly intricate three-dimensional flow field and complex flow properties around the structure, such as nappe interaction, air entrainment, local submergence in inlet keys, and recirculation near sidewalls that depend sensitively on geometric details and discharge events~\cite{Machiels.2012}.
To ensure the safe and efficient operation of PKWs, a comprehensive understanding of the associated hydraulic processes is essential. A central engineering challenge is to predict the discharge efficiency as a function of both geometry and discharge~\cite{luddecke2025key}. Ideally, this capacity must be predicted with sufficient rapidity to facilitate design iteration. However, due to the geometric complexity of the structure, existing studies often focus on a limited number of design configurations, and conclusions as well as empirical design relations may not transfer across a wide range of PKW geometries~\cite{Juestrich.2016, Eslinger.2020, Shen.2021, luddecke2025key}. The complex flow phenomena and the sensitivity of PKW hydraulics to geometric details make discharge performance strongly geometry dependent~\cite{bhukya2022discharge,oertel2016analysis}. \textcolor{black}{From a hydraulic engineering perspective, the prediction of PKW performance remains challenging because the discharge behavior cannot be described by simple reduced relationships comparable to those available for linear weirs. This lack of robust generalized formulations complicates both the hydraulic assessment and the iterative design process of PKWs and highlights the need for approaches that can systematically account for geometric variability.} This sensitivity makes PKWs a challenging testbed for data-driven surrogate models under geometry shifts and out-of-distribution generalization~\cite{elrefaie2024drivaernet++,Tali.2024}.

High-fidelity computational fluid dynamics (CFD) has the theoretical capacity to map performance across a wide spectrum of geometric and discharge ranges. However, large three-dimensional free-surface simulations are computationally intensive, representing a significant challenge in terms of both time and computational resources~\cite{Oertel.2015}. It has been demonstrated that even moderate parametric campaigns can result in the accumulation of substantial CPU hours and multi-terabyte files, with the majority of datasets remaining private and fragmented~\cite{luddecke2025key}. As a result, the civil and hydraulic engineering community lacks the type of openly available, large-scale benchmark datasets that have accelerated progress in other fields and that enable reproducible evaluation of modern machine-learning approaches.

At the same time, engineering research is increasingly advocating the concept of design by data~\cite{ahmed2025design}. Curated datasets that link geometry, operating conditions, and functional performance are viewed as essential infrastructure for learning-based design, surrogate modeling, and inverse design~\cite{ahmed2025design,gebru2021datasheets}. Similar arguments in neighboring domains, such as materials design, emphasize that modern data-centric and representation-learning approaches can help navigate complex physics and uncertainty and thereby accelerate design iteration~\cite{grossmann2023position}. Recent dataset-guideline efforts further stress that engineering datasets should be accompanied by clear documentation of assumptions, modeling and simulation settings, and data provenance~\cite{ahmed2025design,gebru2021datasheets}. They should cover diverse yet physically feasible designs and provide standardized benchmark tasks and splits, together with reusable pipelines under licenses that enable reproduction and extension~\cite{ahmed2025design,gebru2021datasheets}. In aerodynamics and mechanical design~\cite{eilermann20243d}, large-scale datasets such as DrivAerNet~\cite{elrefaie2024drivaernet} and DrivAerNet++~\cite{elrefaie2024drivaernet++} have demonstrated the value of this approach by enabling systematic comparisons across geometric representations, including parameters, point clouds, and meshes, and by evaluating in-distribution and out-of-distribution generalization under explicit distribution-shift protocols. Beyond physics-based surrogates, large language models are increasingly explored as engineering assistants, and recent work highlights the need for rigorous evaluation on real-world engineering tasks~\cite{pradas2024evaluation,heesch2025evaluating}.

Within hydraulic engineering, existing learning-based studies often rely on small tabular datasets derived from experiments, limiting both geometric diversity and reusability~\cite{Anderson.2011,Machiels.2012}. Even when CFD is used, studies frequently vary only one parameter family and typically do not release the underlying geometry and simulation outputs~\cite{Kumar.2024,Laugier.2011}. For PKWs specifically, recent numerical and experimental works provide valuable insights into discharge efficiency and flow characteristics, but they remain confined to restricted geometries or discharge conditions~\cite{luddecke2025key,Shen.2021,shen2023influence}. Related work such as FlowProp demonstrates the promise of data-driven modeling of weir flow properties but is limited to two-dimensional geometries and small sample sizes~\cite{eilermann2024neural}. While generic CFD benchmarks do exist for complex flows, they do not address hydraulic structure design questions, such as discharge coefficients and rating curves across a large variety of geometries~\cite{Tali.2024,Shadkhah.2025,Hassan.2023}.

\textcolor{black}{%
Data-driven surrogate models can enable faster performance estimation in hydraulic engineering design, where conventional CFD pipelines require multiple time-consuming steps. By providing faster predictions, surrogate models can shift slow, CFD-based approaches toward interactive design exploration, allowing hydraulic engineers to iterate quickly and consider a wider range of geometric variations under practical constraints. Recent advances in data-driven surrogate models have demonstrated the potential for predicting accurate estimates at a fraction of the computational cost of high-fidelity CFD \cite{remelli2020meshsdf, kashefi2022physics, song2023surrogate, elrefaie2024drivaernet}. However, many approaches depend on large, high-quality datasets for training and validation, which remain scarce in hydraulic applications. In particular, only a limited number of 3D datasets of PKWs have been reported, typically covering narrow geometric and/or operation condition variations and lacking open public availability \cite{Shen.2021, Bansal.2024, Iqbal.2024, luddecke2025key}. This lack of public, high-quality benchmark datasets prevents robust model training and prevents fair comparison across different approaches, slowing community-driven progress relative to domains that benefited from benchmark datasets \cite{torralba200880, deng2009imagenet, zhou2016thingi10k, wu20153d}.
}

\begin{figure}
\label{Abb_Simulation}
  \includegraphics[width=0.95\linewidth]{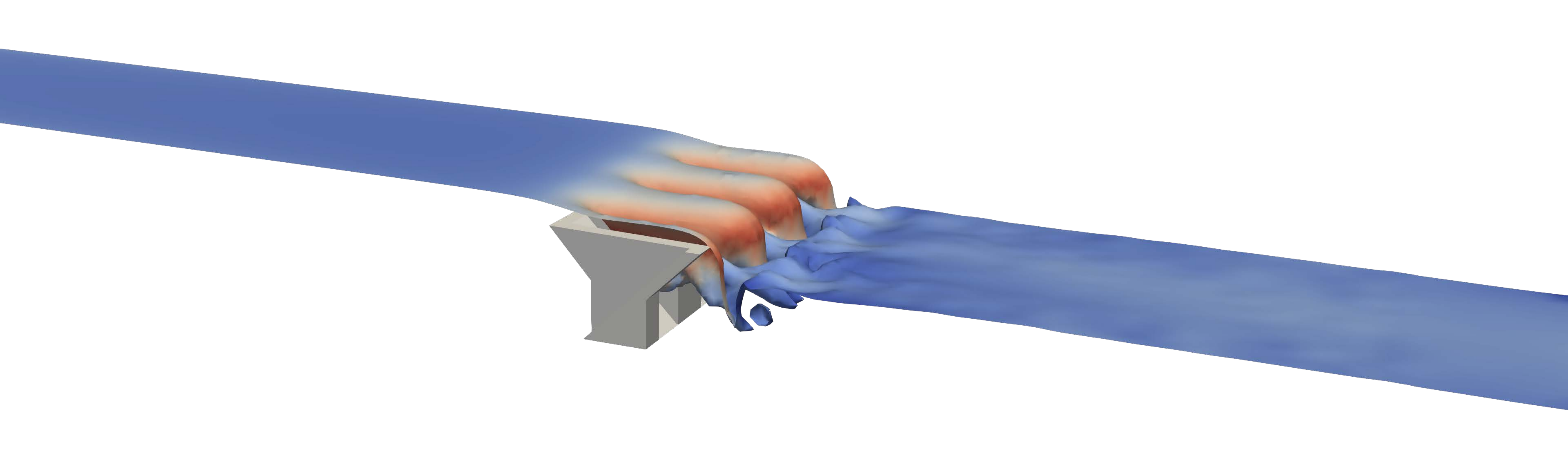}
  \caption{Exemplary results of a CFD simulation of water over a PKW geometry.}
\end{figure}

This work introduces WeirNet, a large-scale 3D CFD benchmark dataset for geometric surrogate modeling of Piano Key Weirs. WeirNet is designed explicitly for AI in civil and hydraulic engineering. It couples a broad, feasibility-constrained PKW design space with free-surface CFD labels and provides multiple geometric modalities that support modern geometric deep learning. Building on and extending the established PKW nomenclature~\cite{pralong2011naming}, a parametric geometry framework is developed that supports both rectangular and trapezoidal plan views and enables automated generation of thousands of structurally feasible designs. For each geometry, a set of 19 free-surface CFD simulations is performed at defined discharges and discharge coefficients are calculated to assess the efficiency of the weir geometries under different discharge events, enabling both forward prediction and curve reconstruction tasks. An exemplary simulation over a Piano Key Weir geometry can be seen in Fig. \ref{Abb_Simulation}.

WeirNet is intended to serve two complementary purposes. First, it supports practical surrogate modeling for rapid screening and design-space exploration, reducing reliance on repeated CFD runs. Second, it provides a research benchmark to study generalization under geometry shifts and operating-condition shifts, and to compare representations (parametric descriptors, point clouds, meshes) and model families under standardized protocols. To encourage reuse and reproducibility, the CAD generation scripts, simulation configurations, post-processing routines, and training pipelines, together with dataset documentation consistent with emerging engineering-dataset recommendations, are released~\cite{ahmed2025design,gebru2021datasheets}.

The key contributions of this work are:
\begin{itemize}
    \item \textbf{Large-scale, multi-modal PKW dataset} A large public 3D PKW benchmark dataset is provided, comprising thousands of parametrically generated geometries, dense CFD simulations per design, and multiple exported representations such as watertight meshes, point clouds, low-dimensional descriptors, and discharge labels.
    \item \textbf{Extended nomenclature and reproducible generation} An extended PKW parameterization is provided that covers trapezoidal plan views and sidewall effects, together with feasibility constraints and a fully reproducible geometry-generation pipeline.
    \item \textbf{Benchmark tasks and baselines} Standardized regression tasks for predicting $c_D$ coefficients and reconstructing rating curves, including out-of-distribution splits that reflect real design scenarios are defined. Strong baselines using parametric ML and geometric deep learning are reported.
\end{itemize}
The WeirNet dataset (CC BY-NC 4.0) and the full training and evaluation code is available at \footnote{Link given after review}. 

\section{Related Work}
\label{sec:sota}

\subsection{Datasets for Civil and Infrastructure Engineering}
In civil and infrastructure engineering, several recent datasets support data-driven design and diagnosis. Chung and Wang~\cite{chung2025dataset} provide a graph-based dataset of complex power-transmission networks for resilient grid design, while Yan et al.\ ~\cite{yan2025bi} release a benchmark dataset for bi-level interturn short-circuit fault monitoring in wind turbine generators. DeepJEB~\cite{hong2025deepjeb} offers a synthetic dataset of jet engine brackets with 3D geometries and structural performance labels for surrogate modeling and inverse design, HUVER~\cite{karri2025huver} collects uncrewed vehicle configurations with aerodynamic and performance data for UAV design, and HM-SYNC~\cite{martins2025hm} captures multimodal human–machine interaction in manufacturing environments. These datasets illustrate how combining geometry or topology with performance metrics and rich metadata enables benchmark tasks for machine learning in engineering design. However, they primarily target structural, energy, or manufacturing systems rather than hydraulic structures.

\subsection{Datasets and Models for Hydraulic and Flow Processes}
Within hydraulic engineering and closely related flow problems, a number of datasets and machine learning studies focus on discharge efficiency prediction and flow fields. Eilermann et al.\cite{eilermann2024neural} showed that Neural Ordinary Differential Equations (NODEs) can accurately learn flow behavior and discharge coefficients $c_D$ from a small set of 2D weir geometries. To support this, they introduced FlowProp, a two-dimensional dataset of weir-type hydraulic structures with flow depths, water-surface profiles, and $c_D$ coefficients~\cite{eilermann2024neural}. More recently, L\"uddecke and Oertel~\cite{luddecke2025key} generated a three-dimensional CFD dataset of trapezoidal PKW geometries with varying key-width ratios, performing $2{,}700$ OpenFOAM\textregistered\ simulations (45 geometries~$\times$~60 discharges) and deriving a polynomial fit for $c_D$ as a function of dimensionless upstream head and key-width ratio. However, this study varies only a single geometric parameter, keeps all other PKW parameters fixed, does not release the dataset publicly, and does not investigate machine-learning surrogates.

In addition, Shen and Oertel~\cite{Shen.2021} compiled a dataset of PKW geometries and discharge coefficients under varying discharges based on laboratory experiments, comparing nonsymmetrical trapezoidal PKWs (TPKWs) and rectangular PKWs (RPKWs) with varying key-width ratios. A broader body of work applies machine-learning models to experimental discharge coefficient data for various weir types, including pseudo-cosine labyrinth weirs~\cite{emami2023lxgb}, trapezoidal arched labyrinth weirs~\cite{heidarnejad2025machine}, and PKWs ~\cite{tian2024enhancing,Bansal.2024,Iqbal.2024}. These studies typically use small tabular datasets (on the order of $10^2$ laboratory observations), focus on scalar $c_D$ prediction from a few dimensionless ratios, and do not release the underlying geometry or CFD fields as reusable, multi-modal datasets.

Beyond PKWs, generic flow benchmarks have been developed. FlowBench~\cite{Tali.2024} is a large-scale dataset of flows over complex two- and three-dimensional obstacles designed to evaluate neural PDE solvers and operator networks. MPF-Bench~\cite{Shadkhah.2025} provides high-fidelity simulations of droplet and bubble dynamics in multiphase flows, while BubbleML~\cite{Hassan.2023} focuses on boiling flows over heated surfaces (nucleate pool boiling, flow boiling, and sub-cooled boiling), with evolving bubbles and liquid–vapor interfaces as the primary “objects” of interest.

Table~\ref{tab:Datasets} summarizes the main CFD- and simulation-based datasets in terms of total size, number of distinct geometries, dimensionality, structure type, geometric variation, available representations, openness, and the primary simulated or measured outputs. The column $\text{SpD}$ (simulations per design) expresses how many operating points (e.g., discharges or boundary-condition settings) are simulated per geometry.
\begin{table}[t]
    \centering
    \caption{Comparison of existing studies and datasets on hydraulic and multiphase flow processes with geometric variation. \emph{Size} counts CFD simulations or experimental cases, $N_{\text{geom}}$ denotes the number of distinct geometries where reported and SpD is the approximate number of simulated operating points (e.g., discharges or boundary conditions) per geometry. P/M/PC indicate availability of parametric descriptors (P), surface or volume meshes (M), and point clouds (PC). The Outputs column summarizes the primary simulated or measured quantities used for surrogate modeling.}
    \label{tab:Datasets}
    \resizebox{\columnwidth}{!}{
    \begin{tabular}{lcccccccc}
        \toprule
        \textbf{Study} &
        \textbf{Size} &
        $N_{\text{geom}}$ &
        \textbf{SpD} &
        \textbf{Dim.} &
        \textbf{Struct.\ type} &
        \textbf{Rep.\ } &
        \textbf{Outputs} &
        \textbf{Open} \\
        & [sims/cases] & [--] & [--] & & &
        (P/M/PC) & & \\
        \midrule
        \textit{Studies on PKW discharge coefficients} & & & & & & & & \\
        \midrule       
        Bansal et al.~\cite{Bansal.2024}&
        162 &
        9 &
        18 &
        3D &
        PKW key width ratio &
        \checkmark / -- / -- &
        $c_D$, Energy dissipation (exp.) &
        -- \\

        Iqbal \& Ghani.~\cite{Iqbal.2024}&
        88 &
        11 &
        8 &
        3D &
        PKW key width ratio \& slope ratio &
        \checkmark / -- / -- &
        $c_D$ (exp.) &
        -- \\        
        \midrule
        \textit{Studies on Weir geometry datasets} & & & & & & & & \\
        \midrule
        
        Eilermann et al.~\cite{eilermann2024neural} &
        15 &
        15 &
        1 &
        2D &
        Weir geometries &
        \checkmark / -- / -- &
        $h(x)$, free surface, $c_D$ &
        \checkmark \\
        
        L\"uddecke \& Oertel~\cite{luddecke2025key} &
        2{,}700 &
        45 &
        60 &
        3D &
        PKW key width ratio (trap.) &
        \checkmark / \checkmark / -- &
        $c_D(H_{t}/P,W_{i}/W_{o})$ (CFD) &
        -- \\

        Shen \& Oertel~\cite{Shen.2021} &
        280 &
        20 &
        14 &
        3D &
        PKW (rect.\ \& trap.\ \& combined) &
        \checkmark / -- / -- &
        $H_{t}/P$-- $c_D$ curves (exp.) &
        -- \\

       Emami et al.~\cite{emami2023lxgb} &
        132 &
        2 &
        66 &
        3D &
        Pseudo‑cosine labyrinth weirs &
        \checkmark / -- / -- &
        $c_D$ (exp.) &
        -- \\

        Heidarnejad et al.~\cite{heidarnejad2025machine} &
        120 &
        12 &
        10 &
        3D &
        Trap.-arched labyrinth weirs &
        \checkmark / -- / -- &
        $c_D$ (exp.) &
        -- \\

        Tian et al.~\cite{tian2024enhancing} &
        476 &
        (/)~*$^{1}$ &
        (/)~*$^{1}$ &
        3D &
        Type-A PKW &
        \checkmark / -- / -- &
        relative discharge $q$ (exp.) &
        -- \\

        \midrule
        \textit{Studies on hydraulic datasets} & & & & & & & & \\
        \midrule
        
        Tali et al.~\cite{Tali.2024} &
        10{,}650 &
        3 sets &
        (/)~*$^{1}$ &
        2D/3D &
        generic bluff bodies &
        -- / \checkmark / -- &
        velocity/pressure fields &
        \checkmark \\
        
        Shadkhah et al.~\cite{Shadkhah.2025} &
        11{,}000 &
        1 &
        11{,}000 &
        2D/3D &
        droplets / bubbles &
        -- / \checkmark / -- &
        multiphase fields (phase, $u,p$) &
        \checkmark \\
        
        Hassan et al.~\cite{Hassan.2023} &
        79 &
        12 &
        1-15 &
        2D/3D &
        boiling surfaces, bubbles &
        -- / \checkmark / -- &
        boiling fields ($T$, phase, $u,p$) &
        \checkmark \\
        
        \textbf{WeirNet (present study)} &
        $\mathbf{71{,}387}$~*$^{2}$ &
        $\mathbf{3{,}794}$ &
        $\mathbf{19}$ &
        3D &
        PKW (rect.\ \& trap.) &
        \checkmark / \checkmark / \checkmark &
        $H_{t}/P$-- $c_D$ curves (CFD) &
        \checkmark \\
        \bottomrule
        \multicolumn{2}{l}{\footnotesize{*1 (not specified)} }\\
        \multicolumn{9}{l}{\footnotesize{*2 (Due to failures and premature terminations in a subset of simulations, a total of 71,387 simulations were completed successfully and constitute the final data set used in this study)} }\\
    \end{tabular}
    }
\end{table}

As Table~\ref{tab:Datasets} shows, existing PKW-related datasets are either two-dimensional, limited to relatively small numbers of three-dimensional geometries, or not publicly accessible. The available 3D PKW datasets vary at most a single geometric ratio (key-width ratio) while keeping all other parameters fixed and provide only mesh-based or purely parametric representations, with scalar $c_D$ or $H_t/P$--$c_D$ curves as outputs. Generic benchmarks such as FlowBench~\cite{Tali.2024}, MPF-Bench~\cite{Shadkhah.2025} and BubbleML~\cite{Hassan.2023} offer substantial geometric and physical diversity and high-fidelity flow fields, but they are not tailored to hydraulic structures or discharge relationships. Experimental ML studies on discharge coefficients for various weir types rely on small, closed tabular datasets without 3D geometry.

In contrast, the WeirNet dataset introduced in this work comprises $3{,}794$ three-dimensional PKW geometries with multiple strongly coupled parameters varied simultaneously (key widths, overhangs, sidewall inclination, sidewall thickness), each simulated at 19 operating points, for a total of $71{,}387$ CFD simulations. For every design, WeirNet provides consistent parametric descriptors (P), watertight surface meshes (M), and high-resolution point clouds (PC), together with discharge coefficients and head-discharge curves. This combination of large geometric diversity, dense sampling in operating conditions, rich geometric modalities, and open availability is not offered by any existing dataset and enables a systematic benchmark for geometric surrogate modeling in hydraulic engineering.

\section{Dataset}
\label{sec:dataset}

\begin{figure}[t]
    \centering
    \includegraphics[width=0.95\linewidth]{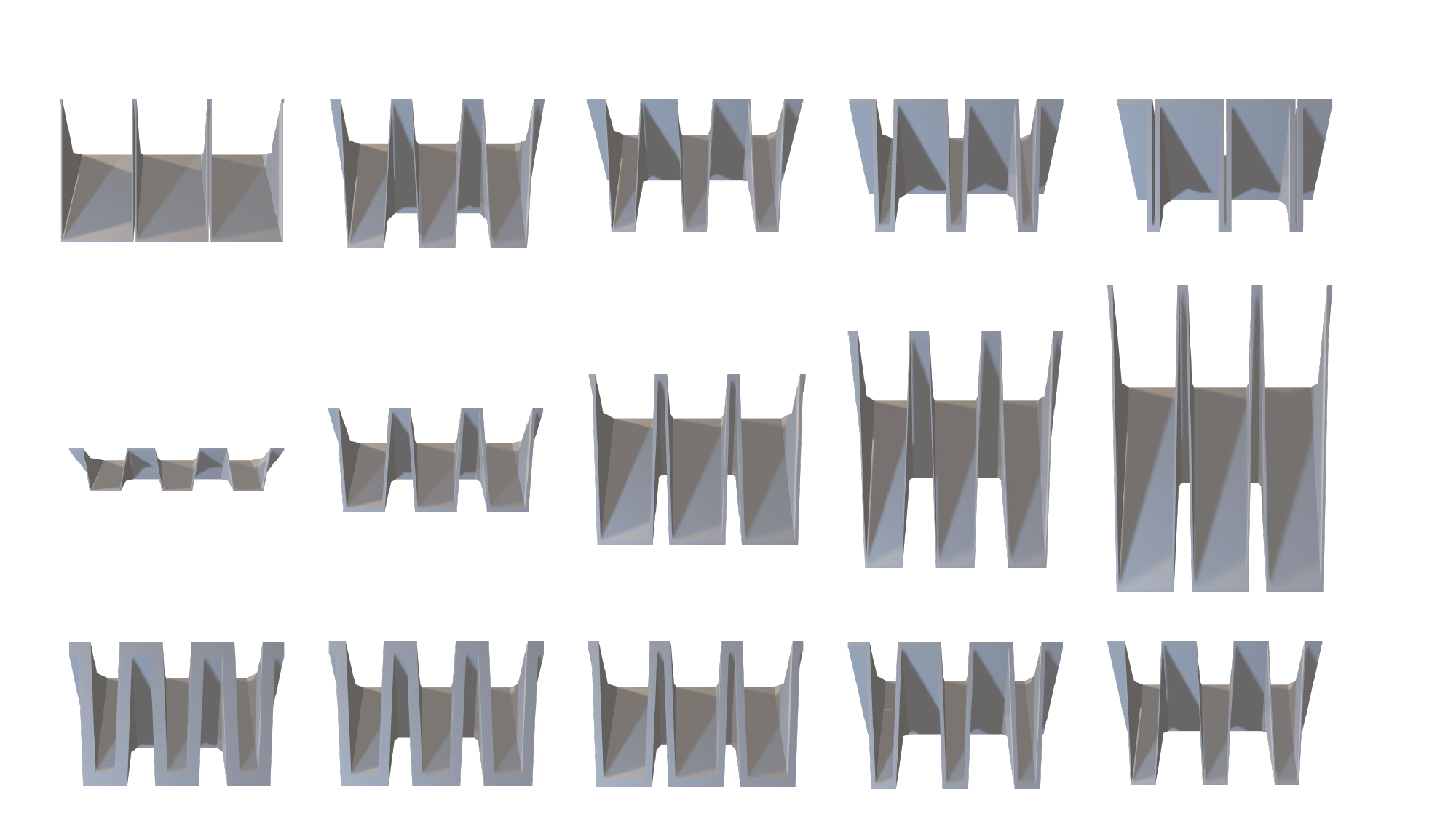}
    \caption{Exemplary geometries from the dataset showing variations in key widths, overhang lengths and wall thickness.}
    \label{fig:STLs}
\end{figure}

\begin{figure}[t]
    \centering
    \includegraphics[width=0.65\linewidth]{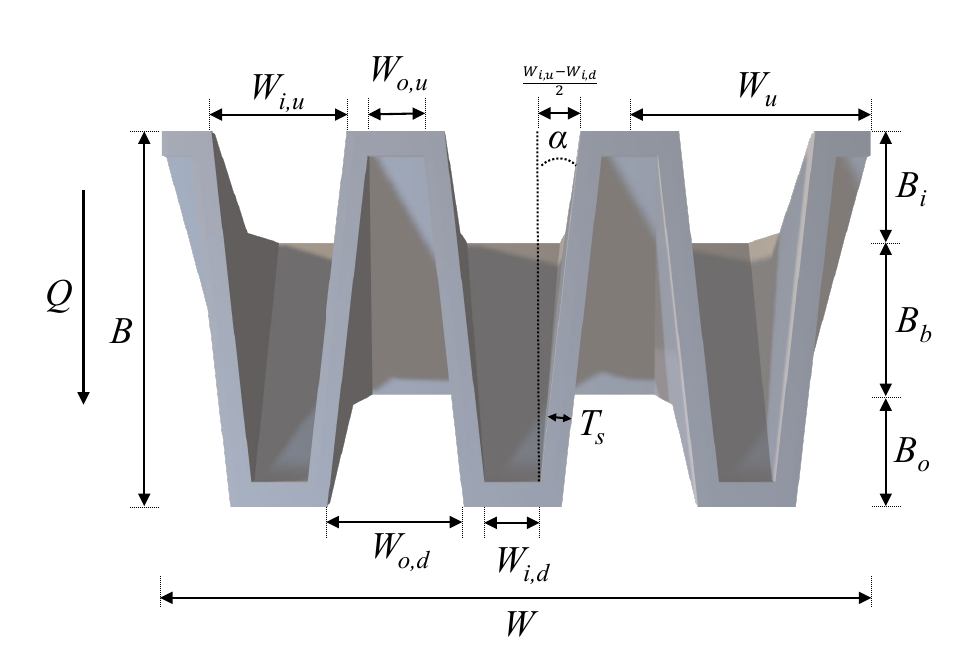}
    \caption{Detailed presentation of geometric relationships between inlet and outlet key widths for TPKWs with $Q$ indicating the direction of flow.}
    \label{Abb_Nomenclature2}
\end{figure}

\paragraph{Geometry Generation}
To successfully develop a parametric model of a PKW, it is necessary to define additional geometric specifications beyond those presented by Pralong et al.~\cite{pralong2011naming}. Furthermore, existing relationships between parameters must be expressed in precise mathematical terms to enable accurate and reproducible model generation. While the nomenclature by Pralong et al.~\cite{pralong2011naming} is based on a RPKW design, improved descriptions and additional definitions of some parameters are necessary to accurately describe other shapes such as the TPKW design. A refined description of the key width definition $W$ has already been widely used in multiple studies. The key width of trapezoidal geometries changes in x-direction, needing definitions of inlet and outlet keys widths ($W_i$ and $W_o$) and upstream and downstream key widths ($W_u$ and $W_d$). A second extension of the existing nomenclature introduces the concept of key length ratios, denoted as $R_{B}$, to describe the overhang lengths of inlet and outlet keys. These ratios relate to the respective key length to the base length $B_b$, providing a dimensionless representation of geometric proportions. The overhang length for each key is therefore resolved as $B_i = R_{B,i} \cdot B_b$ and $B_o = R_{B,o} \cdot B_b$. The updated definitions of key widths are summarized in Table~\ref{TAB_keyWidths} and illustrated in Figure~\ref{Abb_Nomenclature2}. 

\begin{table}[tb]
\caption{Differentiation of the key widths for trapezoidal PKWs and extended nomenclature of the sidewall thickness for trapezoidal PKWs.}
\label{TAB_keyWidths}
\centering
\begin{tabular}{l l}
\toprule
Parameters &	 Definition \\
\midrule
$W_{i,u}$ & Inlet key width (from sidewall to sidewall) on the upstream side \textsuperscript{*1} \\
$W_{i,d}$ & Inlet key width (from sidewall to sidewall) on the downstream side \textsuperscript{*2}\\
$W_{o,u}$ & Outlet key width (from sidewall to sidewall) on the upstream side \textsuperscript{*2} \\
$W_{o,d}$ & Outlet key width (from sidewall to sidewall) on the downstream side \textsuperscript{*1} \\
\midrule
$R_{B,i}$ & Ratio of the downstream (inlet key) overhang crest length to base length \\
$R_{B,o}$ & Ratio of the upstream (outlet key) overhang crest length to base length \\
\midrule
$T_{s}$ & Sidewall thickness - perpendicular to the sidewall \\
$T_{s,2}$ & Sidewall thickness - perpendicular to the main flow direction \\
$T_{s,3}$ & Reduced sidewall thickness - at the outside corner of the inlet and outlet key sidewalls\\
$\Delta T_{s}$ & Difference between $T_{s,2}$ and $T_{s,3}$ \\
\bottomrule
\multicolumn{2}{l}{{\footnotesize *1 (at the beginning of the opening)}}\\
\multicolumn{2}{l}{{\footnotesize *2 (on the inside of the upstream facing crest sidewall)}}\\
\end{tabular}
\end{table}

For a precise representation and mathematical description of trapezoidal PKWs, the conventional parameter of sidewall thickness $T_{s}$ is insufficient. Unlike rectangular PKWs, where the sidewalls are aligned with the flow direction and the thickness remains consistent in the flow-transverse direction, trapezoidal configurations introduce a wall inclination defined by the sidewall angle $\alpha$. This inclination results in an increase of the effective wall thickness perpendicular to the flow direction and it no longer matches the original sidewall thickness $T_{s}$. To account for this, a new parameter $T_{s,2}$ is introduced, representing the projected wall thickness in the flow-transverse direction.
Moreover, an essential aspect requiring explicit definition is the lateral overhang of the sidewall at the outer edge of the weir which exceeds over the predefined key widths. In this study, this overhang is denoted as $T_{s,3}$ and describes a reduced sidewall thickness. The inclusion of $T_{s,3}$ is critical for enabling reliable automated geometry generation in later design stages. Without this parameter, inaccuracies or inconsistencies may arise during geometry construction, particularly at the interfaces between the inclined sidewalls.
The parameters $T_{s}$, $T_{s,2}$ and $T_{s,3}$, along with their geometric interrelations, are illustrated in Figure~\ref{Abb_Nomenclature2}. These additions are especially important due to the demands of working within a fully three-dimensional modeling context. The explicit definition of these parameters allows for geometry generation aligned strictly along the principal axes, without requiring spatial approximations or manual correction in off-axis directions. Omitting these parameters would not only risk producing flawed geometries but would also significantly increase the effort required to ensure geometric accuracy.

The inclination angle $\alpha$ of the sidewall is derived from the difference between the upstream and downstream key widths $W_{i,u}$ and $W_{i,d}$ relative to the total weir length $B$ reduced by the sidewall thickness $T_{s}$ at the inlet and outlet crest. $\alpha$ is calculated as follows:

\begin{align} \label{EQ1}
    \alpha &= \tan^{-1} \cdot \left( \frac{W_{i,u}-W_{i,d}}{2 \cdot \left( B - T_{s} \right)} \right)
\end{align}

Based on this inclination, the effective sidewall thickness in transverse flow direction $T_{s,2}$ can be determined via the cosine relation:
\begin{align} \label{EQ2}
    \cos \left( \alpha \right) = \frac{T_{s}}{T_{s,2}} \Leftrightarrow T_{s,2} = \frac{T_{s}}{\cos \left( \alpha \right)}
\end{align}

To quantify the lateral offset component due to the wall inclination, the projection $\Delta T_{s}$ is introduced. It results from:
\begin{align} \label{EQ3}
    \tan \left( \alpha \right) = \frac{\Delta T_{s}}{T_{s}} \Leftrightarrow \Delta T_{s} = T_{s} \cdot \tan \left( \alpha \right)
\end{align}

Finally, $T_{s,3}$ is calculated as the difference between the inclined wall projection and its lateral offset. Substituting Eq.~\ref{EQ2} and~\ref{EQ3} into this relationship yields the final expression given in Eq.~\ref{EQ5}.
\begin{align} 
    T_{s,3} &= T_{s,2} - \Delta T_{s} \\
    T_{s,3} &= \left( \frac{T_{s}}{\cos \left( \alpha \right)} \right) - \left( T_{s} \cdot \tan \left( \alpha \right) \right) \label{EQ5}
\end{align}
These expressions provide the basis for an accurate and consistent parameterization of the sidewall geometry, enabling robust integration into parametric modeling workflows.

\begin{figure}[t]
    \centering
    \includegraphics[width=0.95\linewidth]{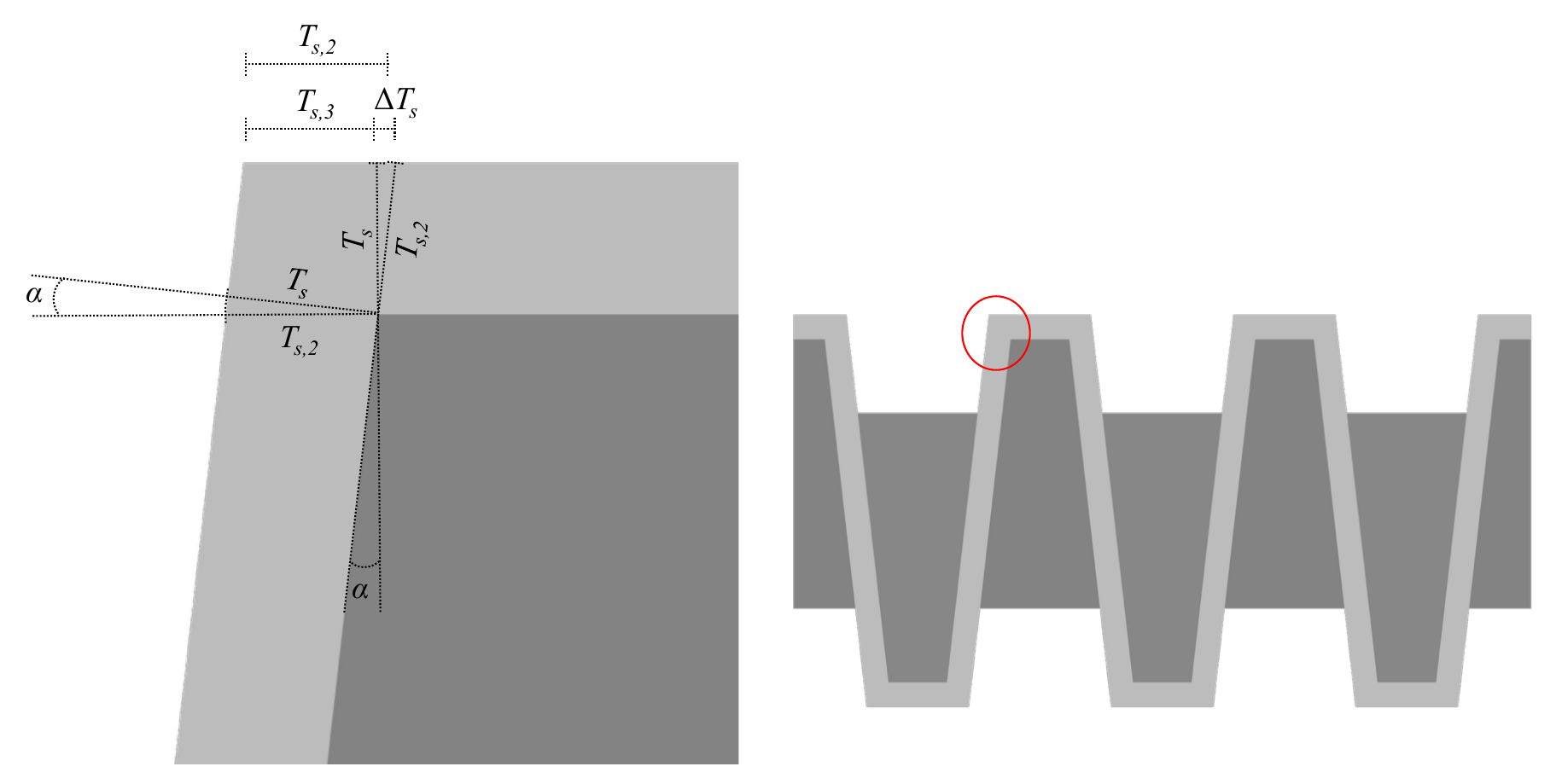}
    \caption{Additional defined parameters at the PKW sidewall. Single parameters are shown on the left, while the corresponding reference locations are indicated on the right.}
    \label{Abb_Nomenclature}
\end{figure}
Designing a PKW involves a comparatively large number of geometric degrees of freedom. Moreover, the admissible parameter set depends on the chosen PKW type (A, B, C, D) and the required crest layout (e.g.~flat or sloped, with or without parapet walls). There are numerous parameters that must be taken into consideration during the design process for Piano Key Weirs. To minimize the dimensionality of the present study \textbf{(i)} all variants are restricted to \emph{PKW Type~A} with a flat crest, \textbf{(ii)} parapet walls, are omitted and \textbf{(iii)} the global height and width of the structure is fixed. Moreover, the variation in PKW geometry was restricted to rectangular and trapezoidal shapes. In addition, all geometries were limited to a three-cycle (3-unit) structure. The resulting model still covers all plan-view aspects that govern hydraulic performance while allowing a single, unified input vector per geometry.
The geometry generation process is constrained by restrictions stemming from interdependencies among various parameters. To avoid parameter settings that result in the generation of non-functioning geometries, pragmatic constraints on geometry input parameters are established. Furthermore, restrictions due to stability and structural feasibility have been considered in the geometrical constraints.
All input parameters were scaled to laboratory conditions using a clear width of 1 meter and a Piano Key Weir height $P$ of 0.33~m. These dimensions were chosen to match an existing physical model~\cite{Besser.2024}, enabling potential comparison of the surrogate model predictions with experimental data.

 The limitations that were selected for the dataset are presented in Table~\ref{TAB_InputParameters}, where $W_{i,u,\mathrm{min}}$ and $W_{i,d,\mathrm{min}}$ refer to the smallest upstream and downstream key widths of 0.03~$\times~P$.
 
 The limitations divide the design parameters into three fixed and five variable design parameters responsible and necessary for the geometry creation. The dataset of geometric designs was generated by using the latin hypercube sampling method (LHS)~\cite{ef76b040-2f28-37ba-b0c4-02ed99573416}  within the limitations mentioned in Table~\ref{TAB_InputParameters} for the five variable parameters. LHS is a highly established method of space filling sampling approaches for the creation of synthetic datasets in the fields of engineering, recommended for example by Picard et. al.~\cite{picard2023dated}. It was chosen because it divides the range of each uncertain variable into equal probability bins, from which samples are randomly selected and matched across different variables to create a sample matrix and enhances the coverage of the uncertain parameter space compared to traditional random sampling techniques~\cite{mohammadi2022efficiency}.

Prior to solid modeling, three analytical constraints are imposed. First, the inlet widths are required to satisfy $W_{i,u} \ge W_{i,d}$, which in turn implies $W_{o,u} \le W_{o,d}$ at the outlet to allow for rectangular geometries ($W_{i,u} = W_{i,d}$ and $W_{o,u} = W_{o,d}$) and trapezoidal geometries ($W_{i,u} > W_{i,d}$ and $W_{o,u} < W_{o,d}$). Second, all geometric widths are restricted to positive values, i.e. $W_{i,u}, W_{i,d}, W_{o,u}, W_{o,d} > 0$. Finally, the global geometry is preserved by keeping the overall height and width constant ($P = \text{const.},\; W = \text{const.}$).

\begin{table}[htb]
\caption{Limitations for the generation of PKW geometries via parametric design.}
\label{TAB_InputParameters}
\centering
\begin{tabular}{l p{2.5 cm} p{3.5 cm}  p{2.5 cm}}
\toprule
Parameters &	 Min   &   Max & Constant \\
\midrule
$W$ & & & 1000 mm\\
$P$ &  & & 330 mm\\
$B_{b}$ & $0.33 \cdot P$ & $1.67 \cdot P$\\
$R_{B,i}$ & 0.25 & 1\\
$N_{u}$ & & & 3\\
$T_{s}$ & $0.015 \cdot P$ & $0.18 \cdot P$\\
$W_{i,u}$ & $0.03 \cdot P$ & $W_{u}-2\cdot T_{s}-W_{i,d,\mathrm{min}}$\\
$W_{i,d}$ & $0.03 \cdot P$ & $W_{u}-2\cdot T_{s}-W_{i,u,\mathrm{min}}$\\
\midrule
    \multicolumn{3}{c}{\emph{Additional constraint:}\;
        $W_{i,u}\;\ge\;W_{i,d}$} \\
\bottomrule
\end{tabular}
\end{table}

\begin{table}[tb]
\caption{Derived parameters in the parametric model}
\label{TAB_DerivedParameters}
\centering
\renewcommand{\arraystretch}{1.25}
\begin{tabular}{l l}
\toprule
\multicolumn{1}{l}{Parameter} & 
\multicolumn{1}{l}{Computational Relationship} \\
\midrule
$W_u$ & $W / N_u$ \\
$B_i$ & $R_{B,i} \cdot B_b$ \\
$B_o$ & $R_{B,o} \cdot B_b$ \\
$B$ & $B_b + B_i + B_o$ \\
$\alpha$ & $\tan^{-1}\left(\frac{W_{i,u} - W_{i,d}}{2 \cdot (B - T_{s})}\right)$ \\
$T_{s,2}$ & $T_{s} / \cos(\alpha)$ \\
$T_{s,3}$ & $T_{s,2} - T_{s} \cdot \tan(\alpha)$ \\
$W_{o,u}$ & $W_u - W_{i,u} - 2 \cdot T_{s,3}$ \\
$W_{o,d}$ & $W_u - W_{i,d} - 2 \cdot T_{s,3}$ \\
\bottomrule
\end{tabular}
\end{table}

All remaining geometric self-intersection checks are delegated to
Rhino’s \texttt{SolidUnion} operation; any failure results in the sample being discarded.

The parametric design methodology was implemented using Rhinoceros 3D (Version 8 SR25) in conjunction with its visual programming plugin, Grasshopper (Version~1.0.0008). This software combination was selected for its capacity to generate complex geometric forms through algorithmic processes and its integration with 3D modeling capabilities. Grasshopper’s node-based visual programming environment enabled the development of parametric models, facilitating iterative design exploration through parameter manipulation. The algorithmic modeling approach allowed for precise geometric control and automated form generation, which proved essential for the systematic investigation of design variations within the research framework. Examples are shown in Fig.~\ref{fig:STLs}. All geometric computations and model generations were executed within this integrated environment, ensuring consistency and reproducibility of the parametric design process. The Grasshopper definition is composed of native components, supplemented by two Python nodes that (i) compute the equations from Table~\ref{TAB_DerivedParameters} and (ii) drive the sampling loop. Continuous ranges are discretized with a 5\,mm step size for lengths and a 0.05 increment for dimensionless factors. With these settings the cartesian product yields \(\sim 22{,}000\) candidate vectors, of which \(\approx 12{,}000\) pass all constraints. All files were extracted in stereolithography format (STL) to enable direct implementation into the CFD software for subsequent investigations.

\paragraph{CFD Simulation}
The numerical CFD model was developed using the open-source CFD software OpenFOAM\textregistered\ v2212. The setup followed established boundary conditions for hydraulic structure simulations as described by Thorenz~\cite{Thorenz.2024}. With a channel width of 1.0 m and a height of 1.0 m the model represents a scaled laboratory system. The computational domain extended 30$~\times~P$ upstream and 15$~\times~P$ downstream of the weir’s upstream crest, resulting in a total length of 15 meters. To achieve a balance between computational efficiency and data accuracy, a base cell size was set to 0.1 m, considering the large number of simulation runs. To capture the geometry of the weir with sufficient precision, the mesh was locally refined twice in the vicinity of the structure. These refinements extended up to 14$~\times~P$ upstream, covering the area from which flow depth data were extracted. To maintain numerical stability and ensure smooth transitions between different cell sizes, single-level transitional refinements were applied between the base mesh and the highly refined zones. The computational grid was generated using blockMesh and snappyHexMesh, resulting in approximately 450,000 to 800,000 cells, depending on the specific weir configuration. Boundary conditions were set according to Thorenz~\cite{Thorenz.2024} with inflow rate at the upstream boundary ($x_{\mathrm{min}}$) and downstream water level at the outflow boundary ($x_{\mathrm{max}}$) serving as key parameters. The downstream flow depth was kept constant at a level that did not affect the overflow behavior at the weir, since downstream flow dynamics were outside the focus of this study. A roughness height of $k_s$~=~0.001 m was specified for the flume bottom ($z_{\mathrm{min}}$), representing a hydraulically smooth boundary. The simulations were performed with the interFOAM solver, which applies the Volume of Fluid (VOF) method to capture free-surface flows. Turbulence was modeled using the Reynolds-Averaged Navier–Stokes (RANS) equations combined with the k-$\omega$ turbulence model. Each simulation was run for 50 seconds to achieve a quasi-steady state. A pressure probe was placed 3$~\times~P$ upstream of the weir crest, recording data every 0.1 seconds. From the recorded pressure time series, flow depths were subsequently derived based on the probe’s specific location. All geometries were simulated under 19 varying discharges, ranging from 50 l/s to 250 l/s, to present a wide range from small discharge events to possible flood events. The exact discharges are listed in Table \ref{TAB_Discharges}.

\begin{table}[tb]
\caption{Simulated discharges $Q$ for all geometries.}
\label{TAB_Discharges}
\centering
\begin{tabular}{l l l l l}
\toprule
\multicolumn{5}{c}{$Q$ [l/s]} \\
\midrule
50 & 80 & 120 & 160 & 200 \\
55 & 90 & 130 & 170 & 225 \\
60 & 100 & 140 & 180 & 250 \\
70 & 110 & 150 & 190 & \\
\bottomrule
\end{tabular}%
\end{table}

\begin{figure}[t]
    \centering
    \includegraphics[width=0.95\linewidth]{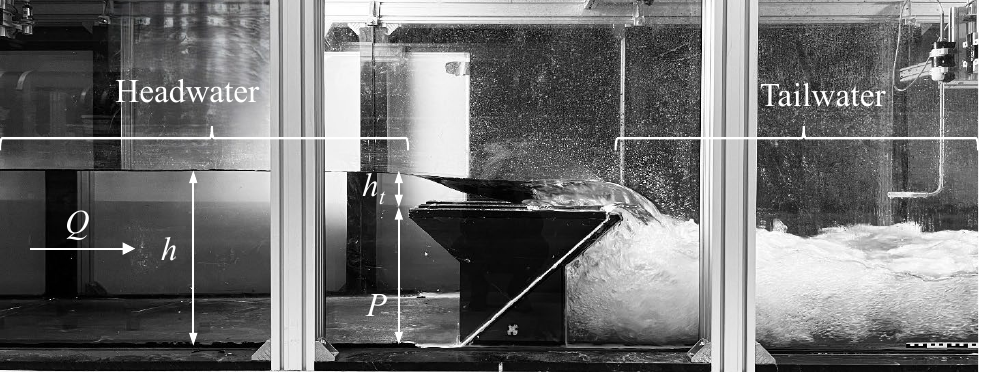}
    \caption{Example of flow behavior over an exemplary PKW geometry in a scaled physical model test with flow direction from left to right.}
    \label{Abb_PKW_Lab}
\end{figure}

\begin{figure}[t]
    \centering
    \includegraphics[width=0.95\linewidth]{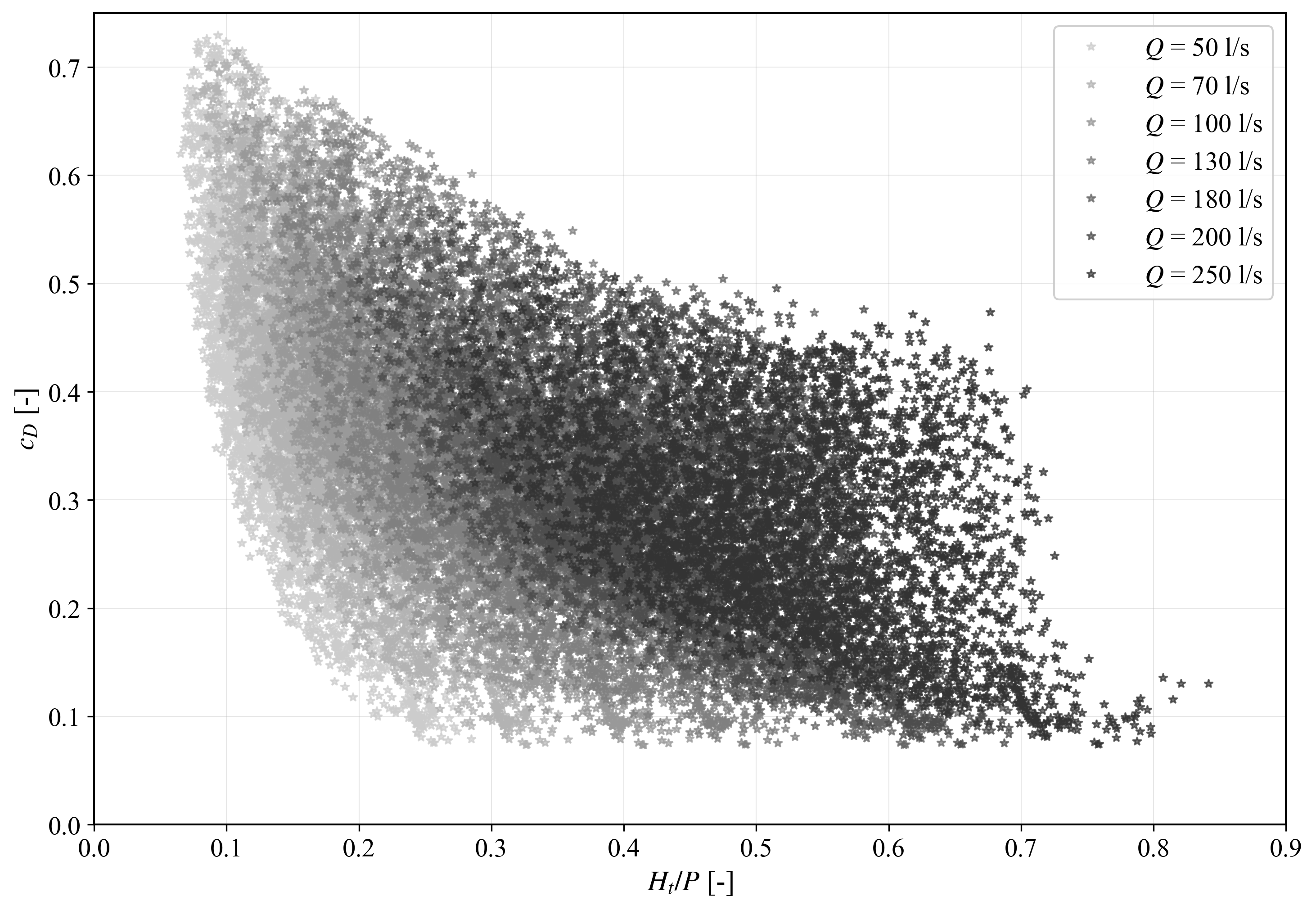}
    \caption{Calculated discharge coefficients from CFD simulations of all investigated geometries for a representative exemplary selection of discharges simulated for the dataset.}
    \label{Abb_Cd}
\end{figure}

To evaluate the data and examine the efficiency of the weir geometries at different discharges, the discharge coefficient $c_D$ was calculated using the Du Buat Formula \cite{Anderson.2011}:

\begin{equation}
\label{eq:discharge}
    Q = \frac{2}{3} \cdot c_D \cdot L \cdot \sqrt{2g} \cdot H_t^{1.5}
\end{equation}
where $Q$ is the discharge, $L$ total developed centerline length along the overflowing crest axis, $g$ is the acceleration due to gravity, and $H_t$ is the total upstream head above the weir crest, calculated as $H_t = v^2/2g + h_t$, in which $v$ is the upstream depth-averaged velocity, and $h_t$ is the flow depth above the weir crest (see Fig. \ref{Abb_PKW_Lab}). Figure~\ref{Abb_Cd} shows calculated discharge coefficients for all tested geometries for a representative selection of discharges.

\begin{figure}[t]
    \centering
    \includegraphics[width=0.95\linewidth]{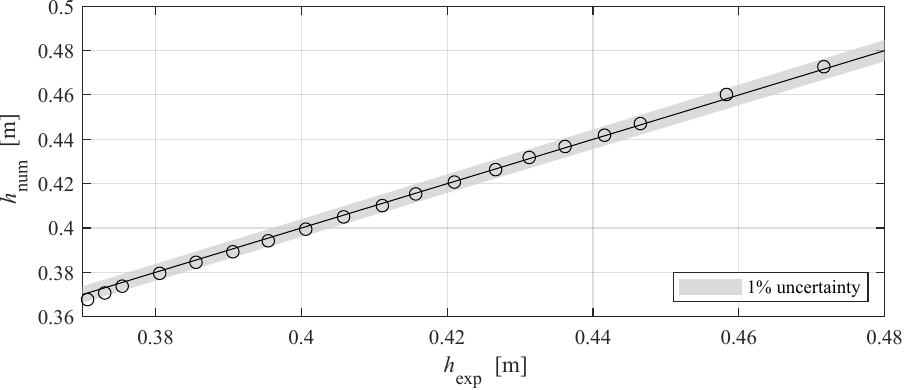}
    \caption{Relative deviation between numerical and experimental results of upstream flow depths used for model validation.}
    \label{Abb_validation}
\end{figure}

Formula \eqref{eq:discharge} rearranged after the $c_D$ follows:
\begin{equation}
\label{eq:discharge_coefficency}
    c_D = \frac{3Q}{2 \, L \, \sqrt{2g} \, H_t^{1{.}5}}
\end{equation}
The numerical model was validated against existing experimental data from a similar setup, differing only in the weir crest geometry. Within the selected range of discharges, the deviations in upstream flow depths between the two datasets were below 1\% (see Fig. \ref{Abb_validation}). 

\paragraph{Computational Cost}

All simulations were performed in parallel computing mode on a high-performance computing (HPC) cluster. In total, 71,387 Simulations were performed running on CPU cores, with a runtime between two and twelve hours, resulting in a raw simulation result file size of 70 TB.

\paragraph{Dataset Structure}
The datasets comprises PKW designs across multiple representations, ensuring compatibility into a variety of applications:
\begin{itemize}
    \item 3D PKW designs as 3D STL meshes
    \item Tabular parametric data of each PKW with 8 geometric parameters
    \item 3D point cloud representations randomly sampled over each PKW surface in $100,000$ points 
    \item CFD results of the PKWs under varying discharge values
    \item CFD data including raw and post-processed data resulting in 70 TB data (not published within publicly available dataset)
\end{itemize}

\textcolor{black}{%
\paragraph{Dataset Access}
This dataset is released under a CC BY-NC 4.0 license.
Full download instructions, metadata, data loader, training, and evaluation code are available at GitHub. Pre-split subsets for training, validation, testing are also provided.
}

\section{Benchmark}
\label{sec:benchmark}
The WeirNet dataset is designed to support reproducible benchmarks for geometric surrogate modeling of hydraulic structures. Building on recent papers on datasets in the fields of engineering design and aerodynamics~\cite{elrefaie2024drivaernet, elrefaie2024drivaernet++, hong2025deepjeb, Tali.2024, Shadkhah.2025}, as well as papers on predicting the discharge coefficient of weir structures~\cite{gharehbaghi2023comparison, belaabed2025optimized, asgharzadeh2025machine}, various machine learning tasks are explored. The focus is on regression-based surrogate modeling of the discharge coefficient $c_D$. 

\paragraph{Dataset Analysis}
\begin{figure}[tb]
    \centering
    \includegraphics[width=0.5\linewidth]{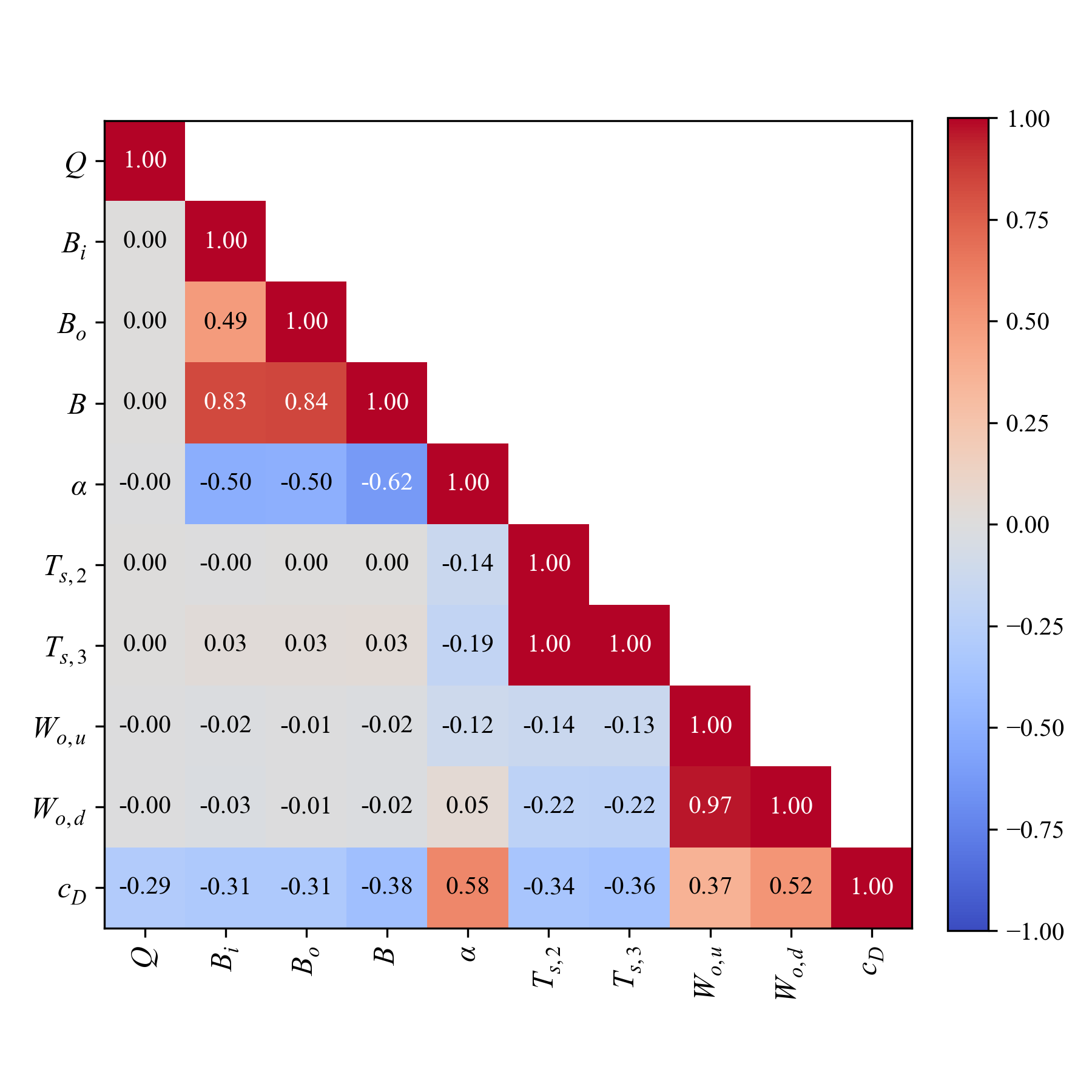}
    \caption{Pearson correlation of selected weir parameters.}
    \label{fig:correlation_heatmap}
\end{figure}

In Fig.~\ref{fig:correlation_heatmap}, the Pearson correlation $\rho$ was computed to quantify linear correlation of selected weir parameters~\cite{benesty2009pearson}. Positive values indicate that the two parameters tend to increase together, with a maximum of $\rho=1$. Negative values indicate inverse correlation, with a minimum of $\rho=-1$. With respect to $c_D$, $\alpha$ and $W_{o,d}$ show a moderate positive correlation, the other parameters only show weak positive or negative correlation ($|\rho|<0.4$)~\cite{papageorgiou2022correlation}. 

\begin{figure}[tb]
    \centering
    \includegraphics[width=\linewidth]{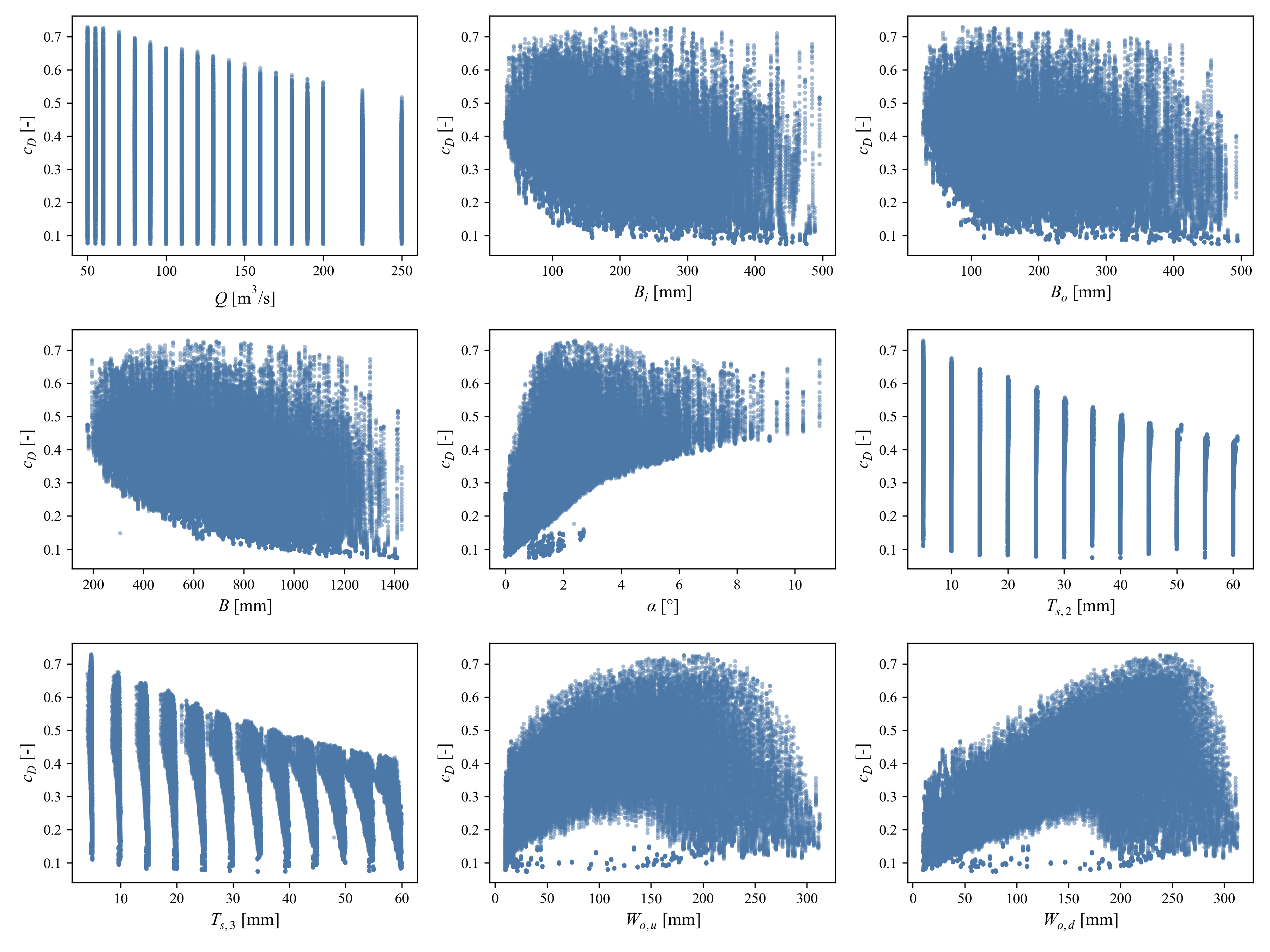}
    \caption{Pairwise correlation between the features and discharge coefficient.}
    \label{fig:correlation_scatter}
\end{figure}

Figure~\ref{fig:correlation_scatter} presents pairwise correlation plots of the discharge coefficient $c_D$ against the discharge $Q$ and selected geometric parameters. Overall, the relationships are weak and predominantly non-linear. $\alpha$ exhibits an increasing trend with $c_D$, consistent with the positive Pearson correlation $\rho$ However, for $\alpha<3^\circ$ the data show increased scatter and several outliers. In contrast, $Q$, $T_{s,2}$, and $T_{s,3}$ display an overall decrease in $c_D$ with increasing values, even though with substantial spread. $B$, $B_i$, and $B_o$ show only a slight negative tendency with considerable scatter. Finally, $W_{o,u}$ and $W_{o,d}$ exhibit a non-linear correlation with $c_D$ showing a curvature behavior.

\subsection{Surrogate Models and Representations}
\label{sec:benchmark_models}
To showcase the multi-modal nature of WeirNet, three representative surrogate models are benchmarked that operate on point clouds and meshes, and four surrogate models that operate on parametric descriptors. 
Similar combinations of parametric and geometric baselines have been used in prior design datasets such as DrivAerNet++~\cite{elrefaie2024drivaernet++}, DeepJEB~\cite{hong2025deepjeb} and linkage-mechanism datasets~\cite{nurizada2025dataset}.

\paragraph{PointNet (point cloud baseline)}
PointNet~\cite{Qi_2017_CVPR} processes unordered 3D point clouds by applying shared multi-layer perceptrons to individual points, followed by global max-pooling to obtain a permutation-invariant shape descriptor. For WeirNet, 5,000 points are randomly sampled from the surface of each PKW mesh, normalized to $[0,1]^3$, concatenated with the scalar discharge $Q$, and the resulting point cloud is fed into PointNet~\cite{Qi_2017_CVPR} to regress $c_D$.

\paragraph{RegDGCNN (dynamic graph on point clouds)}
Regularized Dynamic Graph CNN (RegDGCNN), as used in aerodynamic surrogate modeling for parametric cars~\cite{elrefaie2024drivaernet}, constructs dynamic $k$-nearest-neighbor graphs in feature space and applies edge convolutions to capture local geometric relationships. Following PointNet~\cite{Qi_2017_CVPR} the same point cloud sampling and normalization are used, enabling a direct comparison between global and local point-based architectures. 5,000 points from the surface of each PKW mesh are used, normalized to $[0,1]^3$, concatenated the scalar discharge $Q$, and the resulting point cloud is fed into the model to regress $c_D$.

\paragraph{Mesh-GCN (graph convolution on surface meshes)}
A graph convolutional network (GCN)~\cite{kipf2016semi} operates directly on the triangular surface mesh. Mesh vertices form the nodes of the graph, and mesh edges form the adjacency. Vertex coordinates per weir given from mesh generation in Sec.~\ref{sec:dataset} and $Q$ are used as node features, and several graph-convolution layers aggregate information over the mesh to produce a global geometry descriptor, which is mapped to a scalar prediction of $c_D$. Mesh-based surrogates of this type have shown strong performance in scalar-field prediction on unstructured meshes~\cite{ferguson2024scalar} and aerodynamic surrogate modeling~\cite{elrefaie2024drivaernet++}.

\paragraph{Regression models (parametric data)}
Finally, the PKW geometries as parametric data, are used together with $Q$, to predict $c_D$ using regression models. The scikit-learn framework~\cite{pedregosa2011scikit} is employed using Random Forest~\cite{breiman2001random}, Gradient Boosting~\cite{friedman2001greedy}, XGBoost~\cite{chen2016xgboost} and LightGBM~\cite{ke2017lightgbm} as standalone. The discharge $Q$ and all geometric parameters listed in Table~\ref{TAB_DerivedParameters}, except for the constant parameter $W_u$, are used to predict $c_D$. The study focuses on the eight geometric parameters, which are sufficient to fully define the PKW geometry.

\subsection{Training Setup and Metrics}
\paragraph{Metrics}
\label{sec:benchmark_training} Following prior work~\cite{elrefaie2024drivaernet++,sung2025blendednet}, four standard metrics are reported to evaluate surrogate-model predictions $\hat{c}_{D}$ against CFD discharge coefficients $c_{D}$: Mean Absolute Error $(\mathrm{MAE})$, Maximum Absolute Error $(\mathrm{Max\ AE})$, coefficient of determination $(R^2)$ and Mean Squared Error $(\mathrm{MSE})$. The latter emphasizes larger deviations, $\mathrm{MAE}$ reflects typical error magnitude, $\mathrm{Max\ AE}$ captures worst-case error, and $R^2$ measures explained variance. Overall, lower $\mathrm{MSE}$, $\mathrm{MAE}$, and $\mathrm{Max\ AE}$ and a higher $R^2$ indicate better predictive performance. These metrics are defined as follows:\newline

$\mathrm{MSE}$ averages the squared residuals between CFD discharge coefficients $c_{D}$ and model predictions $\hat{c}_{D}$, penalizing large deviations more strongly.
\begin{equation}
\mathrm{MSE}=\frac{1}{n}\sum_{i=1}^{n}\left(c_{D_i}-\hat{c}_{D_i}\right)^2
\end{equation}

$\mathrm{MAE}$ is the mean absolute residual, summarizing the typical prediction error magnitude and being less dominated by rare large errors than $\mathrm{MSE}$.
\begin{equation}
\mathrm{MAE}=\frac{1}{n}\sum_{i=1}^{n}\left|c_{D_i}-\hat{c}_{D_i}\right|
\end{equation}

$\mathrm{Max\ AE}$ reports the largest absolute residual over the dataset, capturing the worst case discrepancy.
\begin{equation}
\mathrm{Max\ AE}=\max_{i=1}^{n}\left|c_{D_i}-\hat{c}_{D_i}\right|
\end{equation}

$R^2$ measures how much of the variance in $c_d$ is explained by the predictions. Here, $R^2=1$ corresponds to a perfect agreement.
\begin{equation}
R^2 = 1-\frac{\sum_{i=1}^{n}\left(c_{D_i}-\hat{c}_{D_i}\right)^2}{\sum_{i=1}^{n}\left(c_{D_i}-\bar{c}_D\right)^2}
\end{equation} 

Additionally, (i) the number of trainable parameters, (ii) the wall-clock training time per epoch, and (iii) the median inference time for a single sample (batch size 1) on a standard GPU are recorded. These quantities are important for practical deployment in design workflows, where fast evaluation of individual geometries is critical~\cite{elrefaie2024drivaernet++}.

\paragraph{Training Setup}
All models are trained to minimize the $\mathrm{MSE}$ between predicted and true discharge coefficients. For the geometric models, the Adam optimizer~\cite{kingma2017adammethodstochasticoptimization} with a fixed learning rate of $1\times10^{-3}$ and early stopping based on validation loss is used, implemented in PyTorch~\cite{paszke2019pytorch}. Unless stated otherwise, training is performed for up to 500 epochs with a batch size of 32. The XGBoost~\cite{chen2016xgboost}, LightGBM~\cite{ke2017lightgbm}, GradientBoosting~\cite{friedman2001greedy} models are trained using gradient boosting with a learning rate of $5\times10^{-2}$, while the Random Forest~\cite{breiman2001random} models are trained by greedily selecting splits that minimize the MSE.

The dataset is split into 80\% training, 10\% validation, and 10\% test data at the level of geometry–discharge pairs, ensuring that rating curve samples from a given geometry are consistently assigned to the same split. Unless otherwise stated, all results reported in this section use this in-distribution (ID) split. Section~\ref{sec:benchmark_ood} additionally considers out-of-distribution (OOD) splits that hold out subsets of the geometry space and discharge values.

\subsection{Discharge Coefficient Prediction}
\label{sec:benchmark_id}
Table~\ref{tab:results_geo_task1} summarizes the performance of the distribution test on Task~1 for all three models. Point-based methods operate on point clouds with 5,000 points per geometry, while the Mesh-GCN~\cite{kipf2016semi} uses the full surface mesh.
\begin{table}[tb]
    \centering
    \caption{In-distribution surrogate modeling results for predicting $c_D$ at given $Q$ using geometric representation data (Task~1). In addition, the Inference Time for batch size 1 as well as the total number of parameters (\#Params) per model are reported. Reported errors are scaled: $\mathrm{MSE}$ $\times 10^{-5}$, $R^2$ $\times 10^{-2}$, $\mathrm{MAE}$ $\times 10^{-3}$, and $\mathrm{Max\ AE}$ $\times 10^{-1}$.}
    \resizebox{\textwidth}{!}{%
    \begin{tabular}{lccccccc}
        \toprule
        \textbf{Model} &
        $\mathrm{\textbf{MSE (↓)}}$ &
        $\mathbf{R^2}$\textbf{(↑)} &
        $\mathrm{\textbf{MAE (↓)}}$ &
        $\mathrm{\textbf{Max\ AE (↓)}}$ &
        \textbf{Time/epoch (↓)} &
        \textbf{Inference Time (↓)} &
        \textbf{\#Params} \\
        \midrule
        PointNet~\cite{Qi_2017_CVPR} & \textbf{15.30} & \textbf{98.96} & \textbf{9.06} & \textbf{1.67} & \textbf{10\,s} & \textbf{0.49 ms} & 5,435,009 \\
        Mesh-GCN~\cite{kipf2016semi} & 241.60 & 64.33 & 36.88 & 2.99 & 24\,s & 0.90 ms & 100,481 \\
        RegDGCNN~\cite{elrefaie2024drivaernet} & 97.40 & 83.11 & 22.01 & 1.90 & 1054\,s & 53.15 ms & 4,482,657 \\
        \bottomrule
    \end{tabular}%
     }
    \label{tab:results_geo_task1}
\end{table}

\begin{table}[tb]
\centering
\caption{In-distribution surrogate modeling results with regression models for predicting $c_D$ at given $Q$ using parametric data (Task~1). Reported errors are scaled: MSE $\times 10^{-5}$, $R^2$ $\times 10^{-2}$, MAE $\times 10^{-3}$, and Max~MAE $\times 10^{-1}$.}
\begin{tabular}{lcccc}
\toprule
Model & $\mathrm{\textbf{MSE (↓)}}$ & $\mathbf{R^2}$\textbf{(↑)} & $\mathrm{\textbf{MAE (↓)}}$ &
        $\mathrm{\textbf{Max\ AE (↓)}}$ \\ 
\midrule
RandomForest~\cite{breiman2001random}      & \textbf{5.20} & \textbf{99.67} & \textbf{3.94} & 2.90 \\
XGBoost~\cite{chen2016xgboost}           & 6.80 & 99.58 & 5.53 & 1.44 \\
LightGBM~\cite{ke2017lightgbm}          & 7.50 & 99.54 & 6.00 & \textbf{1.05} \\
GradientBoosting~\cite{friedman2001greedy}  & 119.50 & 92.63 & 19.61 & 3.54 \\
\bottomrule
\end{tabular}%
\label{tab:results_param_task1}
\end{table}

\paragraph{Result (geometric)}
The results in Table~\ref{tab:results_geo_task1} show that PointNet~\cite{Qi_2017_CVPR} achieves the best predictive performance among the geometric models, with the lowest $\mathrm{MSE}$ and highest $R^2$, while also being substantially faster to train Mesh-GCN~\cite{kipf2016semi} and RegDGCNN~\cite{elrefaie2024drivaernet}. This aligns with observations in aerodynamic surrogate modeling, where point-based networks often strike a favorable balance between accuracy and computational cost~\cite{elrefaie2024drivaernet}. RegDGCNN~\cite{elrefaie2024drivaernet} attains competitive accuracy with far fewer parameters, but has the highest per-epoch training times because it dynamically reconstructs local graphs in feature space.
Mesh-GCN~\cite{kipf2016semi} performs slightly worse on this task and has a marginally higher per-epoch training time due to graph operations on dense meshes.

In particular, all three geometric surrogates achieve regression error levels that are small compared to typical uncertainties in discharge coefficient measurements and empirical design formulas, suggesting that such models can be used as effective approximations to CFD within PKW design workflows.

\begin{figure}[tb]
    \centering
    \includegraphics[width=0.75\linewidth]{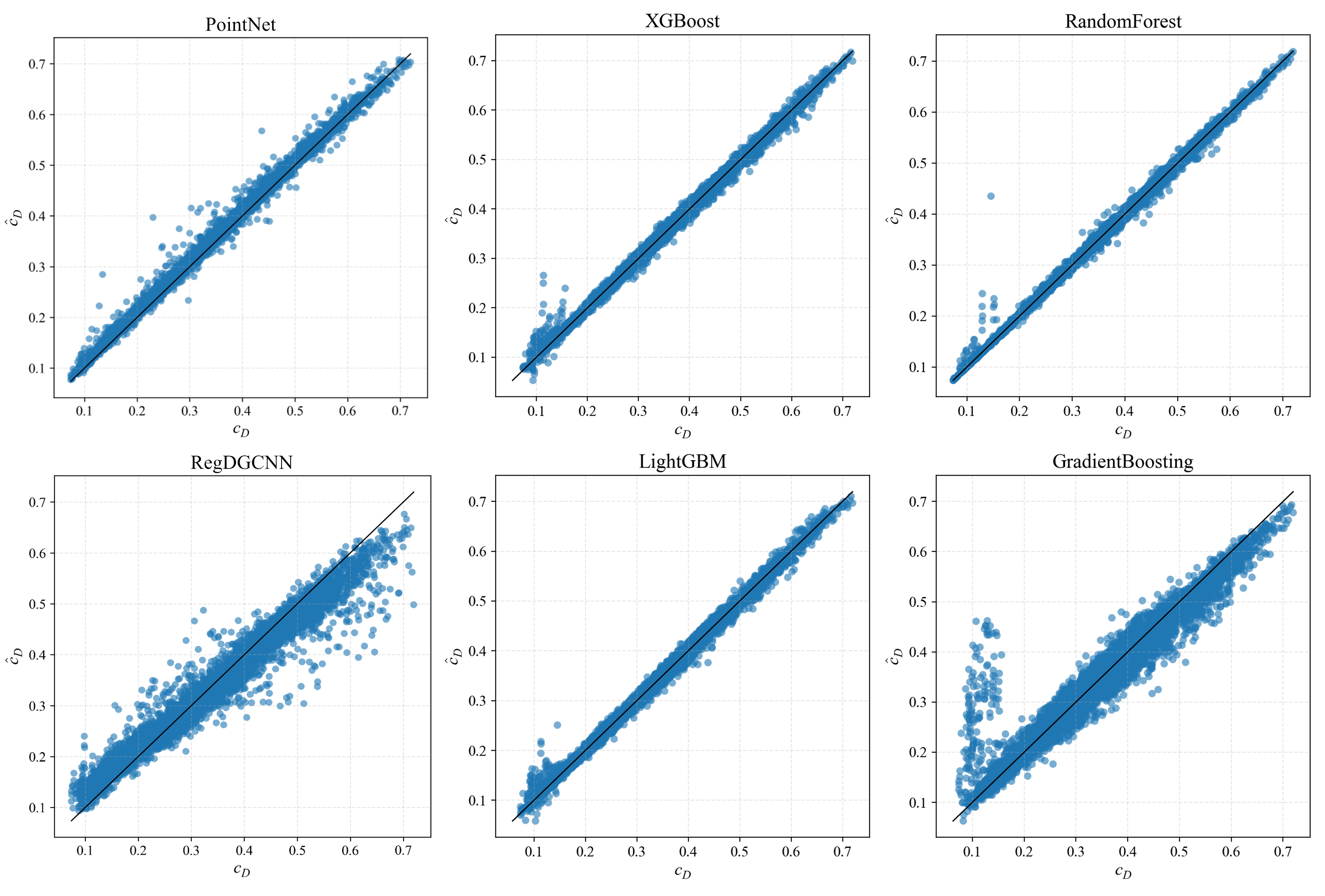}
    \caption{Parity plot of predicted versus CFD discharge coefficients $c_D$ on the in-distribution test set (Task~1).}
    \label{fig:parity_cd}
\end{figure}

\paragraph{Inference Speed}
\label{sec:benchmark_speed}
The inference speed of each geometric surrogate model is evaluated under single-sample inference (batch size 1), which is most relevant for interactive design. For each trained model, the median forward-pass time is measured over 10,000 randomly selected test samples on a standard GPU and CPU. These latencies are reported together with throughput (samples per second) and compared to the typical wall-clock time of a CFD simulation for a single PKW geometry (hours to days; cf. Sec.~\ref{sec:dataset}).

Table~\ref{tab:results_geo_task1} summarizes these measurements. Even with conservative hardware, batch-size-1 inference times are expected to be in the millisecond range, corresponding to a speed-up of several orders of magnitude compared to full CFD. This confirms that WeirNet surrogates can serve as practical proxies for design-space exploration, optimization, and uncertainty studies.

\paragraph{Result (parametric)}
Table~\ref{tab:results_param_task1} summarizes the ID performance for Task 1 of the parametric surrogate models. Overall, Random Forest performs best, achieving the lowest $\mathrm{MSE}$, highest $R^2$, and lowest $\mathrm{MAE}$. Where XGBoost~\cite{chen2016xgboost} and LightGBM~\cite{ke2017lightgbm} achieve slightly less accurate results, with LightGBM~\cite{ke2017lightgbm} providing the best $\mathrm{Max\ AE}$. In contrast, Gradient boosting shows noticeably lower predictive performance.
In addition, Figure~\ref{fig:parity_cd} shows strong prediction performance and confirms the quantitative results.
Only in the smaller $c_D$ area occasional outliers are present, that do not occur for the geometric surrogate models.
Overall, in comparison to the geometric models, all parametric models, besides Gradient Boosting, perform equally well or even better.
It eliminates the need for detailed, complex 3D geometry models, as this geometry can also be described by a few parameters.

\begin{figure}[tb]
    \centering
    \includegraphics[width=1.0\linewidth]{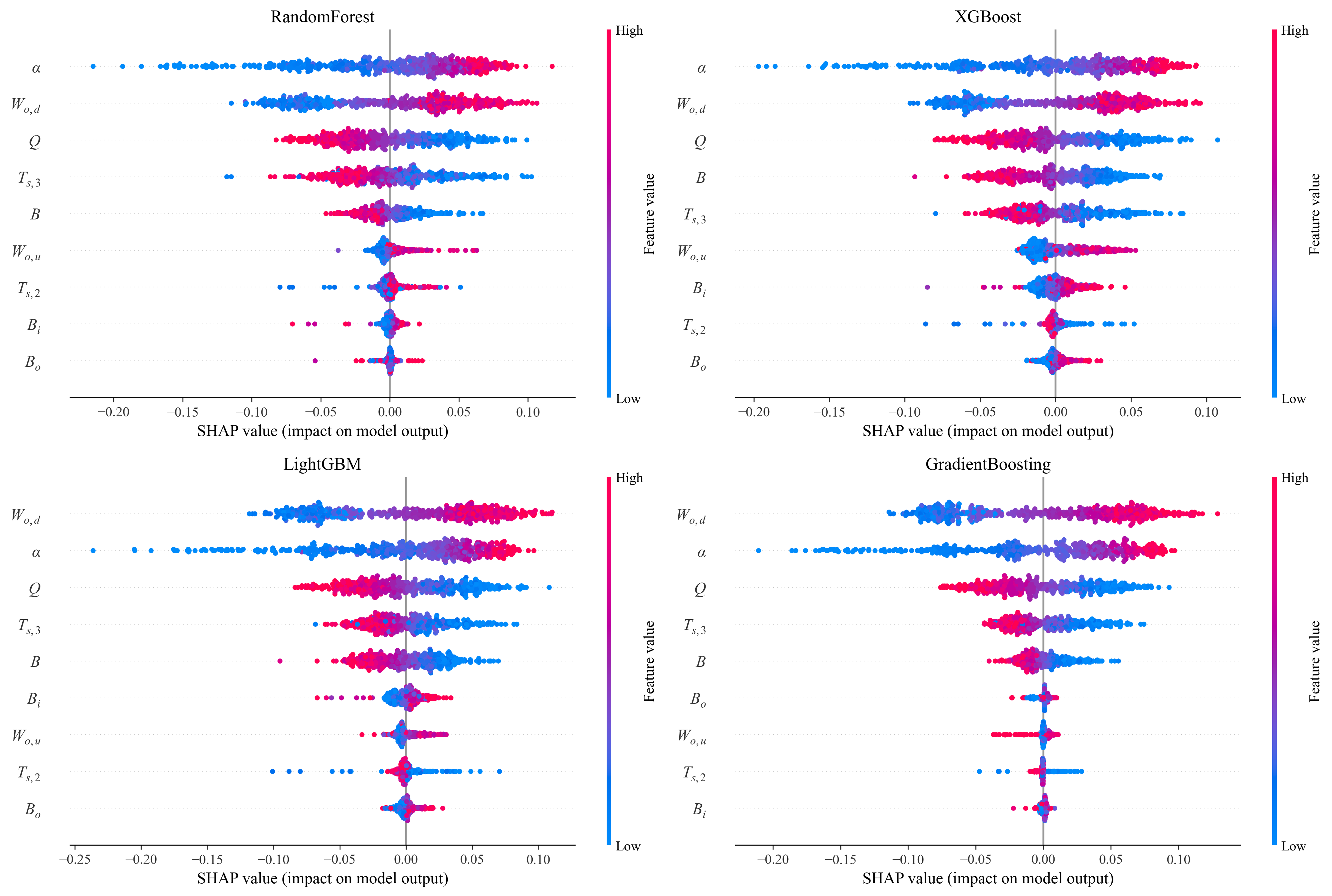}
    \caption{SHAP summary plot for predictions based on all parametric models.}
    \label{fig:shap_summary}
\end{figure}

\paragraph{Models interpretation}
Shapley Additive Explanations (SHAP)~\cite{lundberg2017unified} compute Shapley values~\cite{shapley:book1952} based on principles from game theory to improve the interpretability of model predictions. The Shapley values quantify the contribution of each input feature to the model output. As shown in Figure~\ref{fig:shap_summary}, the SHAP summary plots indicate that higher values of $\alpha$ and $W_{o,d}$ tend to increase $c_D$, whereas lower values tend to decrease it. In contrast, higher values of $Q$, $T_{s,3}$, and $B$ tend to decrease $c_D$, while lower values tend to increase it. The remaining features only have a minor impact on the prediction.

\subsection{Out-of-Distribution Generalization}
\label{sec:benchmark_ood}
\begin{figure}[tb]
    \centering
    \includegraphics[width=\linewidth]{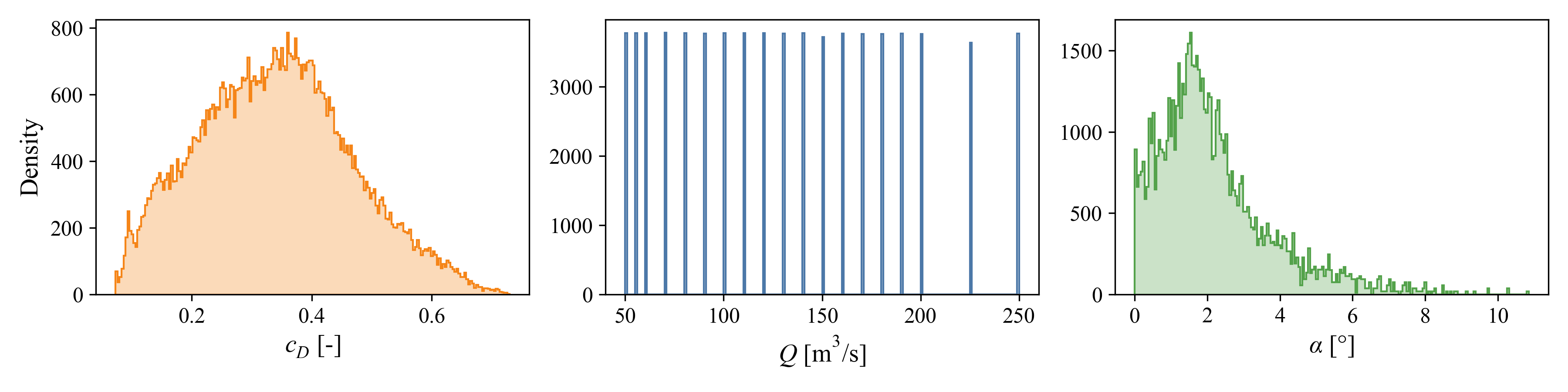}
    \caption{Distribution of the discharge coefficiency $c_D$ (left), discharge $Q$ (middle), and inclination angle $\alpha$ (right).}
    \label{fig:hist}
\end{figure}

To assess the generalization of the parametric surrogate models beyond the training distribution, two OOD splits are defined, inspired by prior designs and benchmarks~\cite{elrefaie2024drivaernet++,Tali.2024,Shadkhah.2025,hong2025deepjeb}.
The splits are based on the distribution shown in Fig.~\ref{fig:hist}, where $c_D$ is an approximate Gaussian distribution, $Q$ is a discrete distribution, and $\alpha$ is a right skewed distribution.

\paragraph{OOD-Geom}
To evaluate generalization across PKW geometries, three geometric splits based on the sidewall inclination angle $\alpha$ are defined to train and test the models on each split separately. Specifically, (i) models are trained on steeper sidewall inclinations and tested on shallower or rectangular PKWs ($\alpha_{\leq 2^\circ}$), (ii) trained on shallower and steeper inclinations and tested on moderate inclinations ($\alpha_{3^\circ\text{--}5^\circ}$), and (iii) trained on rectangular and shallower PKWs and tested on steeper inclinations ($\alpha_{\ge 6^\circ}$).

\paragraph{OOD-Head}
Extrapolation across unseen discharge values for fixed PKWs is evaluated using three splits: (i) models trained on the upper low third of the head range and tested on the lower third ($Q_{\leq 90}$), (ii) trained on the lowest and highest thirds and tested on the middle third ($Q_{100\text{--}160}$), and (iii) trained on the lowest two thirds and tested on the highest third ($Q_{\ge 170}$).

\paragraph{Results}
For each split, $\mathrm{MSE}$, $R^2$, $\mathrm{MAE}$, and $\mathrm{Max\ AE}$ are reported in Table~\ref{tab:ood_results}.
The results show that generalizing towards new discharge values are considerably easier than generalizing to new geometries.
For OOD-Head, performance remains close to ID behaviour across all discharge ranges, with slight degradation observed during unseen lower ($Q_{\leq 90}$) and higher ($Q_{\ge 170}$) discharge conditions. This suggests that, once a PKW family has been represented during training, surrogate models can predict unseen discharges with little loss of fidelity.
In contrast, OOD-Geom exposes that $c_D$ cannot be predicted by training them on trapezoidal PKWs and testing them on rectangular PKWs or trapezoidal PKWs with smaller sidewall inclination ($\alpha_{\leq 2^\circ}$) due to non-linear behavior, see Figure~\ref{fig:correlation_scatter}. Despite that, the performance is strong for the mid and high inclination ranges ($\alpha_{3^\circ\text{--}5^\circ}$, and $\alpha_{\ge 6^\circ}$).
Overall, these results indicate that dataset diversity in geometry is the bottleneck for OOD robustness, whereas a wider range of discharge values is expected to provide smaller improvements.

\begin{table}[tb]
    \centering
    \caption{Out-of-distribution performance ($\mathrm{MSE}$, $R^2$, $\mathrm{MAE}$, and $\mathrm{Max\ AE}$) for all surrogate models on in-distribution (ID), OOD-Geom, and OOD-Head splits. Reported errors are scaled: MSE $\times 10^{-5}$, $R^2$ $\times 10^{-2}$, MAE $\times 10^{-3}$, and Max~MAE $\times 10^{-1}$.}
    \label{tab:ood_results}
    \begin{tabular}{lccccc}
        \toprule
        \textbf{Model} & \textbf{Split} & $\mathrm{\textbf{MSE (↓)}}$ & $\mathbf{R^2}$ \textbf{(↑)}& $\mathrm{\textbf{MAE (↓)}}$ &  \makecell{$\mathrm{\textbf{Max}}$\\$\mathrm{\textbf{MAE (↓)}}$}\\
        \midrule
        RandomForest~\cite{breiman2001random}      & $\alpha_{\leq 2^\circ}$           & 1628.40 & -19.43 & 94.34 & 5.27 \\
        XGBoost~\cite{chen2016xgboost}             & $\alpha_{\leq 2^\circ}$           & 1350.50 & 0.95   & 86.13 & \textbf{4.78} \\
        LightGBM~\cite{ke2017lightgbm}             & $\alpha_{\leq 2^\circ}$           & \textbf{1242.40} & \textbf{8.88}   & \textbf{83.35} & 4.80 \\
        GradientBoosting~\cite{friedman2001greedy} & $\alpha_{\leq 2^\circ}$           & 1582.60 & -16.07 & 94.28 & 5.19 \\
        \midrule
        RandomForest~\cite{breiman2001random}      & $\alpha_{3^\circ\text{--}5^\circ}$ & 87.20   & \textbf{87.59}  & 19.11 & 4.93 \\
        XGBoost~\cite{chen2016xgboost}             & $\alpha_{3^\circ\text{--}5^\circ}$ & \textbf{37.10}   & 94.72  & \textbf{14.53} & 1.57 \\
        LightGBM~\cite{ke2017lightgbm}             & $\alpha_{3^\circ\text{--}5^\circ}$ & 38.60   & 94.51  & 15.37 & 1.36 \\
        GradientBoosting~\cite{friedman2001greedy} & $\alpha_{3^\circ\text{--}5^\circ}$ & 61.20   & 91.28  & 19.77 & \textbf{1.02} \\
        \midrule
        RandomForest~\cite{breiman2001random}      & $\alpha_{\ge 6^\circ}$            & 63.70   & 84.84  & 20.07 & 1.01 \\
        XGBoost~\cite{chen2016xgboost}             & $\alpha_{\ge 6^\circ}$            & 42.70   & 89.83  & 16.54 & \textbf{0.67} \\
        LightGBM~\cite{ke2017lightgbm}             & $\alpha_{\ge 6^\circ}$            & \textbf{41.50}   & 90.13  & \textbf{16.14} & 0.76 \\
        GradientBoosting~\cite{friedman2001greedy} & $\alpha_{\ge 6^\circ}$            & 119.60  & \textbf{71.54}  & 29.14 & 1.11 \\
        \midrule
        RandomForest~\cite{breiman2001random}      & $Q_{\leq 90}$                     & 138.20  & 92.66  & 28.90 & 1.84 \\
        XGBoost~\cite{chen2016xgboost}             & $Q_{\leq 90}$                     & \textbf{135.00}  & 92.82  & \textbf{28.41} & \textbf{1.79} \\
        LightGBM~\cite{ke2017lightgbm}             & $Q_{\leq 90}$                     & 146.60  & 92.21  & 29.64 & 1.85 \\
        GradientBoosting~\cite{friedman2001greedy} & $Q_{\leq 90}$                     & 315.40  & \textbf{83.23}  & 40.03 & 3.61 \\
        \midrule
        RandomForest~\cite{breiman2001random}      & $Q_{100\text{--}160}$             & 55.80   & 96.32  & 17.52 & 1.95 \\
        XGBoost~\cite{chen2016xgboost}             & $Q_{100\text{--}160}$             & 144.80  & 90.45  & 29.27 & 1.91 \\
        LightGBM~\cite{ke2017lightgbm}             & $Q_{100\text{--}160}$             & \textbf{51.60}   & 96.59  & \textbf{16.97} & \textbf{1.30} \\
        GradientBoosting~\cite{friedman2001greedy} & $Q_{100\text{--}160}$             & 161.00  & \textbf{89.38}  & 25.15 & 3.40 \\
        \midrule
        RandomForest~\cite{breiman2001random}      & $Q_{\ge 170}$                     & 94.90   & 91.02  & 21.61 & 1.69 \\
        XGBoost~\cite{chen2016xgboost}             & $Q_{\ge 170}$                     & \textbf{92.00}   & 91.30  & \textbf{20.96} & \textbf{1.66} \\
        LightGBM~\cite{ke2017lightgbm}             & $Q_{\ge 170}$                     & 92.60   & 91.25  & 21.50 & 1.68 \\
        GradientBoosting~\cite{friedman2001greedy} & $Q_{\ge 170}$                     & 186.30  & \textbf{82.38}  & 30.27 & 2.75 \\
        \bottomrule
    \end{tabular}
\end{table}

\subsection{Data Efficiency}
\label{sec:benchmark_dataeff}
Because CFD simulations of PKWs are computationally expensive (Sec.~\ref{sec:dataset}), it is important to understand how much data is required to obtain useful surrogates. Following DrivAerNet++~\cite{elrefaie2024drivaernet++} and DeepJEB~\cite{hong2025deepjeb}, PointNet~\cite{Qi_2017_CVPR} and all parametric models are trained on increasing fractions $\{10\%, 20\%, 40\%, 60\%, 80\%, 100\%\}$ of the training set, keeping the validation and test sets fixed.

For each model and fraction, the $R^2$ on the test set are reported. 
The resulting learning curves illustrate how additional CFD simulations translate into surrogate accuracy and can guide practitioners in deciding when further data collection yields diminishing returns.

\begin{figure}[tb]
    \centering
    \includegraphics[width=0.8\linewidth]{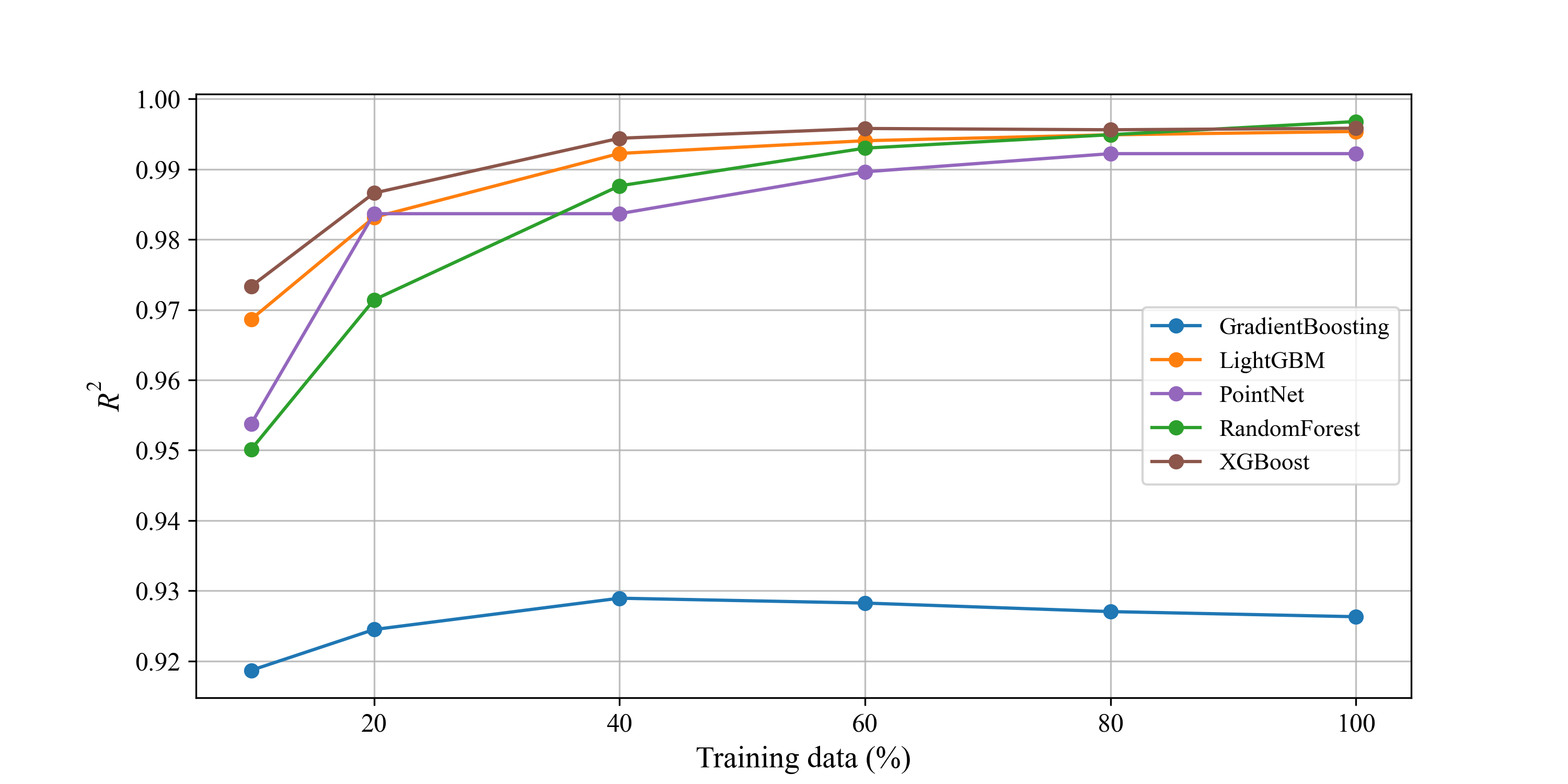}
    \caption{Data-efficiency of surrogate models on WeirNet.}
    \label{fig:data_efficiency}
\end{figure}

\paragraph{Result}
Figure~\ref{fig:data_efficiency} shows that for most models, the $R^2$ improves as the training data size increases. Beyond $60\%$ training data, gains become marginal, plateauing from $80\%$ to $100\%$. In contrast, gradient boosting shows degraded performance with additional training data, indicating limited capacity as a surrogate model.

\section{Discussion}
\label{sec:discussion}
The benchmark experiments in Sec.~\ref{sec:benchmark} indicate that comparatively simple surrogate models can approximate CFD-derived discharge coefficients for PKWs with good accuracy. The observed error magnitudes are comparable to, and in several cases smaller than, those reported for data-driven discharge-coefficient prediction based on small experimental datasets in hydraulic engineering~\cite{emami2023lxgb,heidarnejad2025machine,tian2024enhancing}. Among the tested parametric surrogates, tree-based regression models trained on the extended PKW parameterization, together with discharge $Q$, attain the best in-distribution performance while remaining lightweight and fast at inference. This supports the notion that the extended nomenclature and geometric model introduced in Sec.~\ref{sec:dataset} compress the relevant design variability into a compact descriptor set, consistent with observations from engineering design datasets where carefully chosen parametric features can rival more complex geometric encodings for scalar targets~\cite{elrefaie2024drivaernet,elrefaie2024drivaernet++,hong2025deepjeb}.

In Sec.~\ref{sec:benchmark_id}, among the geometric deep-learning models on point clouds and meshes, only PointNet~\cite{Qi_2017_CVPR} remains competitive on the in-distribution split. RegDGCNN~\cite{elrefaie2024drivaernet} performs moderately well in comparison, whereas Mesh-GCN~\cite{kipf2016semi} clearly falls behind, at the cost of higher computational overhead. Their main advantage lies in their representational flexibility, as they operate directly on 3D geometry and do not depend on a specific CAD parameterization. This becomes particularly relevant for future extensions of WeirNet beyond these Type of PKWs or for applications where only mesh data are available. Similar trade-offs between parametric and geometric models have been reported in DrivAerNet++~\cite{elrefaie2024drivaernet++}, DeepJEB~\cite{hong2025deepjeb}, and related benchmarks~\cite{nurizada2025dataset}. This suggests that hybrid workflows using parametric surrogates where possible and geometric models as a fallback are a pragmatic choice in engineering design.

The out-of-distribution splits in Sec.~\ref{sec:benchmark_ood} and Table~\ref{tab:ood_results} demonstrate that high in-distribution accuracy alone is not sufficient for safe deployment. For the OOD-Head split, where models are evaluated on discharge ranges not seen during training while geometries remain within the training distribution, performance degrades only moderately and remains usable for preliminary design screening. In contrast, the OOD-Geom split based on held-out ranges of sidewall inclination $\alpha$ causes substantially larger performance drops. This indicates that geometry shifts are the dominant failure mode in this setting, consistent with broader evidence from geometric surrogate benchmarks that distribution shifts in shape space are more challenging than shifts in operating conditions~\cite{Tali.2024,Shadkhah.2025,elrefaie2024drivaernet++}. The comparatively stronger behavior of physically meaningful descriptors and global point-based representations suggests that emphasizing global structure can mitigate geometry extrapolation to some degree. Nevertheless, the remaining gap between in-distribution and OOD performance is non-negligible and should motivate uncertainty-aware deployment or fallback to CFD for out-of-manifold designs.

The data-efficiency study in Sec.~\ref{sec:benchmark_dataeff} provides quantitative support for the data-driven design perspective. Surrogate accuracy improves rapidly as more simulated samples are added, before exhibiting diminishing returns beyond roughly $60\%$ of the training data in the setup. Interpreting these fractions at the geometry level suggests that a few hundred geometries with full rating curves can already yield strong predictive performance, while the full scale of WeirNet remains valuable for closing residual error and enabling more robust evaluation under challenging splits.

\textcolor{black}{Overall, the results suggest that data-driven surrogate models can reliably support the early stages of PKW design by enabling rapid comparison of geometries and operating conditions, while high-fidelity CFD and physical experiments remain indispensable for final verification and for configurations outside the explored design space. The strong in-distribution performance suggests that the models effectively capture the predominant hydraulic factors regulating discharge efficiency and can therefore support rapid comparison of design alternatives, and preliminary rating curve estimation. At the same time, the observed degradation under geometric extrapolation reflects the well-known sensitivity of PKW hydraulics to shape variations and underscores the continued need for high-fidelity CFD and physical experiments in final verification and for designs outside the explored parameter space. The principal advantage of the surrogate approach lies in its computational efficiency. Predictions can be obtained in fractions of a second, whereas a single three-dimensional free-surface CFD simulation may require hours of computation, and physical experiments involve substantial time, resources, and setup effort. This gain in computational speed allows a much broader range of geometries and discharge scenarios to be assessed during the planning phase than would be feasible using CFD or physical modeling alone, thereby improving the basis for hydraulic design decisions.}

\paragraph{Broader impact}
By providing an open, large-scale 3D CFD benchmark with reproducible geometry and simulation pipelines, WeirNet can accelerate CFD-driven weir design by enabling surrogate models that replace many expensive trial simulations with millisecond-level predictions for screening and optimization. The standardized tasks, splits, and baselines facilitate method comparison and diagnosis of failure modes (e.g., under geometry shifts), which can improve reliability prior to deployment. Reducing the number of required CFD runs can additionally lower computational energy use. More broadly, WeirNet creates a shared testbed connecting hydraulic engineering, geometric deep learning, and scientific computing.

\paragraph{Limitations}
WeirNet is intentionally scoped and therefore its conclusions should be interpreted within the dataset’s domain of validity. First, the benchmark targets a restricted PKW design space with fixed global dimensions and excludes other PKW types as well as additional geometric features such as parapet walls or noses. Next, all labels are derived from a single CFD workflow, so residual model form and numerical errors may propagate into the learned surrogates and limit transfer to alternative CFD setups or experimental measurements. Finally, the dataset focuses on scalar performance targets (e.g., $c_D$ coefficients) rather than full flow fields, uncertainty estimates or multi-objective criteria. Since all simulations were performed at laboratory scale uses a single RANS $k$--$\omega$ turbulence model, the results may be affected by scaling effects.

\section{Conclusion and Outlook}
\label{sec:conclusion}
WeirNet contributes a large-scale 3D CFD dataset for PKWs that helps close the gap between generic flow benchmarks and the small, typically closed datasets used in hydraulic structure surrogate modeling. Building on an extended PKW nomenclature and a reproducible parametric geometry-generation framework, WeirNet comprises 3,794  trapezoidal designs simulated at 19 discharges, and provides multiple geometric representations together with discharge coefficients and rating curves. Combined with the standardized tasks, splits, and baselines in Sec.~\ref{sec:benchmark}, this establishes a reusable framework for geometric surrogate modeling in civil and hydraulic engineering.

The benchmark results show that parametric surrogates trained on the extended descriptor set can reach or exceed the accuracy of more complex geometric deep-learning models for scalar $c_D$ prediction, while being substantially cheaper to train and deploy. This underlines the value of domain knowledge and carefully constructed parameterizations in engineering machine learning, complementing observations from other engineering design benchmarks~\cite{elrefaie2024drivaernet,elrefaie2024drivaernet++,hong2025deepjeb}. At the same time, point- and mesh-based models remain important when CAD parameters are unavailable, when the design space cannot be fully captured by a small set of scalars, or when future tasks require richer targets than scalar performance curves. Overall, learned surrogates reduce per-query evaluation from expensive CFD runs to millisecond-scale inference on modern hardware, enabling interactive exploration and optimization of PKW geometries.

\textcolor{black}{WeirNet demonstrates that data-driven surrogate models can complement established numerical and experimental approaches by enabling rapid estimation of discharge coefficients across a broad range of geometries and operating conditions. While CFD and physical modeling remain essential for final verification and detailed analysis, the presented dataset shows that much of the early-stage exploration of PKW designs can be shifted toward fast surrogate-based evaluation, supporting coefficient-based hydraulic design and analysis. This capability allows engineers to investigate a substantially wider design space during planning than would be feasible with traditional methods alone.}

Looking ahead, WeirNet offers several avenues for future research. Extending the design space to additional PKW types, crest layouts, and varying global dimensions would stress-test the relative merits of parametric and geometric surrogates and move closer to project-scale scenarios. Incorporating multi-fidelity simulations and targeted laboratory measurements would enable calibration and multi-fidelity learning, aligning with current discussions on dataset quality and documentation in engineering design~\cite{ahmed2025design,picard2023dated,rad2024datasets,gebru2021datasheets}. Methodologically, WeirNet can support studies of uncertainty quantification under geometric and operating-condition shifts, benchmark inverse and multi-objective design algorithms, and train conditional generative models for hydraulic structure design~\cite{hohmann2024design,eilermann20243d,elrefaie2024drivaernet,elrefaie2024drivaernet++,hong2025deepjeb}. Recent point cloud generative backbones that could be explored in this context include flow-based models~\cite{yang2019pointflow,wu2023fast}, diffusion models for point clouds~\cite{luo2021diffusion}, latent point diffusion~\cite{vahdat2022lion} and continuous-time consistency models enabling few-step sampling~\cite{eilermann2025continuous}.

Further on, the raw data set of CFD simulations (70 TB of raw data and result files) could be used as a base data set to restart simulations with increased mesh fidelity or more complex turbulence models, which could be used to create training data sets for more complex hydraulic research questions.  

By releasing not only the data but also CFD configurations and training pipelines this work supports reproducibility and extendability beyond a single research group. WeirNet is anticipated to provide both a practical tool for hydraulic engineers in weir design and a benchmark for the geometric deep-learning and scientific machine-learning communities, supporting advances toward reliable data-driven design tools for hydraulic infrastructure. 

\section*{Acknowledgments}
M. Hohmann is funded by the Deutsche Forschungsgemeinschaft (DFG, German Research Foundation) – 520460697
S. Eilermann is member of project LaiLa which is funded by dtec.bw – Digitalization and Technology Research Center of the Bundeswehr, which we gratefully acknowledge. dtec.bw is funded by the European Union – NextGenerationEU.
Computational resources (HPC cluster HSUper) have been provided by the project hpc.bw, funded by dtec.bw – Digitalization and Technology Research Center of the Bundeswehr. 


\bibliographystyle{plain}  
\bibliography{references}

@inproceedings{eilermann2024neural,
  title={A Neural Ordinary Differential Equations Approach for 2D Flow Properties Analysis of Hydraulic Structures},
  author={Eilermann, Sebastian and L{\"u}ddecke, Lisa and Hohmann, Michael and Zimmering, Bernd and Oertel, Mario and Niggemann, Oliver},
  booktitle={1st ECAI Workshop on “Machine Learning Meets Differential Equations: From Theory to Applications”},
  pages={1--17},
  year={2024},
  organization={PMLR}
}

@incollection{luddecke2025key,
  title={Key width ratio influence on discharge coefficients of trapezoidal Piano Key Weirs via openFOAM simulations},
  author={L{\"u}ddecke, Lisa and Oertel, Mario},
  booktitle={Labyrinth and Piano Key Weirs IV},
  pages={130--137},
  year={2025},
  publisher={CRC Press}
}

@article{Eslinger.2020,
  title={Energy dissipation of type a piano key weirs},
  author={Eslinger, Kam R. and Crookston, Brian M.},
  journal={Water},
  volume={12},
  number={5},
  pages={1253},
  year={2020},
  publisher={MDPI}
}

@article{Juestrich.2016,
  title={Mobile riverbed scour downstream of a piano key weir},
  author={J{\"u}strich, Stefan and Pfister, Michael and Schleiss, Anton J},
  journal={Journal of Hydraulic Engineering},
  volume={142},
  number={11},
  pages={04016043},
  year={2016},
  publisher={American Society of Civil Engineers}
}

@article{LeiteRibeiro.2012,
  title={Discharge capacity of piano key weirs},
  author={Leite Ribeiro, Marcelo and Bieri, M and Boillat, J-L and Schleiss, AJ and Singhal, G and Sharma, N},
  journal={Journal of Hydraulic engineering},
  volume={138},
  number={2},
  pages={199--203},
  year={2012},
  publisher={American Society of Civil Engineers}
}

@article{Lemperiere.2003,
  title={The Piano Keys weir: a new cost-effective solution for spillways},
  author={Lemp{\'e}ri{\`e}re, Fran\c{c}ois and Ouamane, Ahmed},
  journal={International Journal on Hydropower \& Dams},
  volume={10},
  number={5},
  pages={144--149},
  year={2003}
}

@book{Machiels.2012,
  title={Experimental study of the hydraulic behaviour of Piano Key Weirs},
  author={Machiels, Olivier},
  year={2012},
  publisher={Universite de Liege (Belgium)}
}

@article{pralong2011naming,
  title={A naming convention for the piano key weirs geometrical parameters},
  author={Julien Pralong and Julien Vermeulen and Benoit Blancher and Fr{\'e}d{\'e}ric Laugier and S{\'e}bastien Erpicum and Olivier Machiels and Michel Pirotton and J. L. Boillat and Marcelo Leite Ribeiro and Anton J. Schleiss},
  journal={Labyrinth and piano key weirs},
  volume={271},
  pages={278},
  year={2011},
  publisher={Taylor \& Francis London}
}

@inproceedings{Schleiss.2011,
  title={From labyrinth to piano key weirs: A historical review},
  author={Schleiss, Anton J.},
  booktitle={Proc. Int. Conf. Labyrinth and Piano Key Weirs Li{\`e}ge B},
  pages={3--15},
  year={2011}
}

@article{shen2023influence,
  title={Influence of piano key weir crest shapes on flow characteristics, scale effects, and energy dissipation for in-channel application},
  author={Shen, Xiaoyang and Oertel, Mario},
  journal={Journal of Hydraulic Engineering},
  volume={149},
  number={6},
  pages={04023010},
  year={2023},
  publisher={American Society of Civil Engineers}
}

@article{Shen.2021,
  title={Comparative study of nonsymmetrical trapezoidal and rectangular piano key weirs with varying key width ratios},
  author={Shen, Xiaoyang and Oertel, Mario},
  journal={Journal of Hydraulic Engineering},
  volume={147},
  number={11},
  pages={04021045},
  year={2021},
  publisher={American Society of Civil Engineers}
}

@article{Tali.2024,
  title={Flowbench: A large scale benchmark for flow simulation over complex geometries},
  author={Tali, Ronak and Rabeh, Ali and Yang, Cheng-Hau and Shadkhah, Mehdi and Karki, Samundra and Upadhyaya, Abhisek and Dhakshinamoorthy, Suriya and Saadati, Marjan and Sarkar, Soumik and Krishnamurthy, Adarsh and others},
  journal={arXiv preprint arXiv:2409.18032},
  year={2024}
}

@article{Shadkhah.2025,
  title={MPFBench: A Large Scale Dataset for SciML of Multi-Phase-Flows: Droplet and Bubble Dynamics},
  author={Shadkhah, Mehdi and Tali, Ronak and Rabeh, Ali and Yang, Cheng-Hau and Herron, Ethan and Upadhyaya, Abhisek and Krishnamurthy, Adarsh and Hegde, Chinmay and Balu, Aditya and Ganapathysubramanian, Baskar},
  journal={arXiv preprint arXiv:2502.07080},
  year={2025}
}

@article{Hassan.2023,
  title={Bubbleml: A multiphase multiphysics dataset and benchmarks for machine learning},
  author={Hassan, Sheikh Md Shakeel and Feeney, Arthur and Dhruv, Akash and Kim, Jihoon and Suh, Youngjoon and Ryu, Jaiyoung and Won, Yoonjin and Chandramowlishwaran, Aparna},
  journal={Advances in Neural Information Processing Systems},
  volume={36},
  pages={418--449},
  year={2023}
}

@inproceedings{Thorenz.2024,
  title={Boundary conditions for hydraulic structures modelling with OpenFOAM},
  author={Thorenz, Carsten},
  booktitle={Proceedings of the 10th International Symposium on Hydraulic Structures (ISHS 2024)},
  pages={76--86},
  year={2024},
  organization={ETH Zurich}
}

@InProceedings{Qi_2017_CVPR,
author = {Qi, Charles R. and Su, Hao and Mo, Kaichun and Guibas, Leonidas J.},
title = {PointNet: Deep Learning on Point Sets for 3D Classification and Segmentation},
booktitle = {Proceedings of the IEEE Conference on Computer Vision and Pattern Recognition (CVPR)},
month = {July},
year = {2017}
}

@inproceedings{kipf2016semi,
  author = {Kipf, Thomas N. and Welling, Max},
  booktitle = {ICLR (Poster)},
  ee = {https://openreview.net/forum?id=SJU4ayYgl},
  publisher = {OpenReview.net},
  title = {Semi-Supervised Classification with Graph Convolutional Networks.},
  url = {http://dblp.uni-trier.de/db/conf/iclr/iclr2017.html#KipfW17},
  year = 2017
}

@misc{kingma2017adammethodstochasticoptimization,
      title={Adam: A Method for Stochastic Optimization}, 
      author={Diederik P. Kingma and Jimmy Ba},
      year={2017},
      eprint={1412.6980},
      archivePrefix={arXiv},
      primaryClass={cs.LG},
      url={https://arxiv.org/abs/1412.6980}, 
}

@inproceedings{elrefaie2024drivaernet,
  title={Drivaernet: A parametric car dataset for data-driven aerodynamic design and graph-based drag prediction},
  author={Elrefaie, Mohamed and Dai, Angela and Ahmed, Faez},
  booktitle={International Design Engineering Technical Conferences and Computers and Information in Engineering Conference},
  volume={88360},
  pages={V03AT03A019},
  year={2024},
  organization={American Society of Mechanical Engineers}
}

@article{elrefaie2024drivaernet++,
  title={Drivaernet++: A large-scale multimodal car dataset with computational fluid dynamics simulations and deep learning benchmarks},
  author={Elrefaie, Mohamed and Morar, Florin and Dai, Angela and Ahmed, Faez},
  journal={Advances in Neural Information Processing Systems},
  volume={37},
  pages={499--536},
  year={2024}
}

@inproceedings{sung2025blendednet,
  title={Blendednet: A blended wing body aircraft dataset and surrogate model for aerodynamic predictions},
  author={Sung, Nicholas and Spreizer, Steven and Elrefaie, Mohamed and Samuel, Kaira and Jones, Matthew C and Ahmed, Faez},
  booktitle={International Design Engineering Technical Conferences and Computers and Information in Engineering Conference},
  volume={89237},
  pages={V03BT03A049},
  year={2025},
  organization={American Society of Mechanical Engineers}
}

@article{ahmed2025design,
  title={Design by Data: Cultivating Datasets for Engineering Design},
  author={Ahmed, Faez and Picard, Cyril and Chen, Wei and McComb, Christopher and Wang, Pingfeng and Lee, Ikjin and Stankovic, Tino and Allaire, Douglas and Menzel, Stefan},
  journal={Journal of Mechanical Design},
  volume={147},
  number={4},
  pages={040301},
  year={2025},
  publisher={American Society of Mechanical Engineers}
}

@article{ferguson2024scalar,
  title={Scalar field prediction on meshes using interpolated multiresolution convolutional neural networks},
  author={Ferguson, Kevin and Gillman, Andrew and Hardin, James and Kara, Levent Burak},
  journal={Journal of Applied Mechanics},
  volume={91},
  number={10},
  pages={101002},
  year={2024},
  publisher={American Society of Mechanical Engineers}
}

@inproceedings{picard2023dated,
  title={Dated: Guidelines for creating synthetic datasets for engineering design applications},
  author={Picard, Cyril and Schiffmann, J{\"u}rg and Ahmed, Faez},
  booktitle={International Design Engineering Technical Conferences and Computers and Information in Engineering Conference},
  volume={87301},
  pages={V03AT03A015},
  year={2023},
  organization={American Society of Mechanical Engineers}
}

@article{rad2024datasets,
  title={Datasets in design research: needs and challenges and the role of AI and GPT in filling the gaps},
  author={Rad, Mohammad Arjomandi and Hajali, Tina and Bonde, Julian Martinsson and Panarotto, Massimo and W{\"a}rmefjord, Kristina and Malmqvist, Johan and Isaksson, Ola},
  journal={Proceedings of the Design Society},
  volume={4},
  pages={1919--1928},
  year={2024},
  publisher={Cambridge University Press}
}

@article{Bansal.2024,
  title={Prediction of inlet-to-outlet width ratio of type-A piano key weir using fuzzy neural network (FNN)},
  author={Bansal, Nipun and Bhardwaj, Keshav and Singh, Deepak and Kumar, Munendra},
  journal={Journal of Water and Climate Change},
  volume={15},
  number={9},
  pages={4368--4388},
  year={2024},
  publisher={IWA Publishing}
}

@article{Iqbal.2024,
  title={Prediction of the discharge capacity of piano key weirs using artificial neural networks},
  author={Iqbal, Mujahid and Ghani, Usman},
  journal={Journal of Hydroinformatics},
  volume={26},
  number={5},
  pages={1167--1188},
  year={2024},
  publisher={IWA Publishing}
}

@article{Kumar.2024,
  title={Numerical and statistical analysis of auxiliary geometrical parameter effects on piano key weir discharge capacity},
  author={Kumar, Binit and Ahmad, Rahil and Pandey, Manish and Gupta, Anil Kumar},
  journal={AIMS Environmental Science},
  volume={11},
  number={5},
  pages={723--740},
  year={2024}
}

@inproceedings{Anderson.2011,
  title={Influence of Piano Key Weir geometry on discharge},
  author={Anderson, Richard M and Tullis, Blake},
  booktitle={Proceedings of the international conference labyrinth and piano key weirs},
  pages={75--80},
  year={2011}
}

@inproceedings{Besser.2024,
  title={Tailwater influence on downstream flow conditions of piano key weirs},
  author={Besser, Lisa and Oertel, Mario},
  booktitle={Proceedings of the 10th International Symposium on Hydraulic Structures (ISHS 2024)},
  pages={336--345},
  year={2024},
  organization={ETH Zurich}
}

@inproceedings{Laugier.2011,
  title={Influence of structural thickness of sidewalls on PKW spillway discharge capacity},
  author={Laugier, Frederic and Pralong, Julien and Blancher, Benoit},
  booktitle={Proc. Intl Workshop on Labyrinths and Piano Key Weirs PKW},
  pages={159--165},
  year={2011}
}

@article{gebru2021datasheets,
  title={Datasheets for datasets},
  author={Gebru, Timnit and Morgenstern, Jamie and Vecchione, Briana and Vaughan, Jennifer Wortman and Wallach, Hanna and Iii, Hal Daum{\'e} and Crawford, Kate},
  journal={Communications of the ACM},
  volume={64},
  number={12},
  pages={86--92},
  year={2021},
  publisher={ACM New York, NY, USA}
}

@article{hong2025deepjeb,
  title={Deepjeb: 3d deep learning-based synthetic jet engine bracket dataset},
  author={Hong, Seongjun and Kwon, Yongmin and Shin, Dongju and Park, Jangseop and Kang, Namwoo},
  journal={Journal of Mechanical Design},
  volume={147},
  number={4},
  pages={041703},
  year={2025},
  publisher={American Society of Mechanical Engineers}
}

@article{nurizada2025dataset,
  title={A dataset of 3M single-DOF planar 4-, 6-, and 8-bar linkage mechanisms with open and closed coupler curves for machine learning-driven path synthesis},
  author={Nurizada, Anar and Dhaipule, Rohit and Lyu, Zhijie and Purwar, Anurag},
  journal={Journal of Mechanical Design},
  volume={147},
  number={4},
  pages={041702},
  year={2025},
  publisher={American Society of Mechanical Engineers}
}

@article{karri2025huver,
  title={HUVER: The HyForm Uncrewed Vehicle Engineering Repository},
  author={Karri, Abhiram and Stump, Gary and McComb, Christopher and Song, Binyang},
  journal={Journal of Mechanical Design},
  volume={147},
  number={4},
  pages={044505},
  year={2025},
  publisher={American Society of Mechanical Engineers}
}

@article{chung2025dataset,
  title={Dataset on Complex Power Systems: Design for Resilient Transmission Networks Using a Generative Model},
  author={Chung, In-Bum and Wang, Pingfeng},
  journal={Journal of Mechanical Design},
  volume={147},
  number={4},
  pages={041709},
  year={2025},
  publisher={American Society of Mechanical Engineers}
}

@article{yan2025bi,
  title={Bi-Level Interturn Short-Circuit Fault Monitoring for Wind Turbine Generators With Benchmark Dataset Development},
  author={Yan, Jingyi and Senemmar, Soroush and Zhang, Jie},
  journal={Journal of Mechanical Design},
  volume={147},
  number={4},
  pages={041704},
  year={2025},
  publisher={American Society of Mechanical Engineers}
}

@article{martins2025hm,
  title={HM-SYNC: A Multimodal Dataset of Human Interactions With Advanced Manufacturing Machinery},
  author={Martins, John and Lin, Cheyu and Flanigan, Katherine A and McComb, Christopher},
  journal={Journal of Mechanical Design},
  volume={147},
  number={4},
  pages={044504},
  year={2025},
  publisher={American Society of Mechanical Engineers}
}

@article{grossmann2023position,
  title={Position Paper on Materials Design--A Modern Approach},
  author={Grossmann, Willi and Eilermann, Sebastian and Rensmeyer, Tim and Liebert, Artur and Hohmann, Michael and Wittke, Christian and Niggemann, Oliver},
  journal={arXiv preprint arXiv:2312.10996},
  year={2023}
}

@article{paszke2019pytorch,
  title={Pytorch: An imperative style, high-performance deep learning library},
  author={Paszke, Adam and Gross, Sam and Massa, Francisco and Lerer, Adam and Bradbury, James and Chanan, Gregory and Killeen, Trevor and Lin, Zeming and Gimelshein, Natalia and Antiga, Luca and others},
  journal={Advances in neural information processing systems},
  volume={32},
  year={2019}
}

@article{bhukya2022discharge,
  title={Discharge estimation over piano key weirs: A review of recent developments},
  author={Bhukya, Raj Kumar and Pandey, Manish and Valyrakis, Manousos and Michalis, Panagiotis},
  journal={Water},
  volume={14},
  number={19},
  pages={3029},
  year={2022},
  publisher={MDPI}
}

@inproceedings{oertel2016analysis,
  title={Analysis of various piano key weir geometries concerning discharge coefficient development},
  author={Oertel, Mario and Bremer, Frederik. L.},
  booktitle={Proceedings of the 4th IAHR Europe Congress, Liege, Belgium},
  pages={27--29},
  year={2016}
}

@inproceedings{hohmann2024design,
  title={Design automation: a conditional VAE approach to 3D object generation under conditions},
  author={Hohmann, Michael and Eilermann, Sebastian and Gro{\ss}mann, Willi and Niggemann, Oliver},
  booktitle={2024 IEEE 29th International Conference on Emerging Technologies and Factory Automation (ETFA)},
  pages={1--8},
  year={2024},
  organization={IEEE}
}

@article{eilermann20243d,
  title={3d multi-criteria design generation and optimization of an engine mount for an unmanned air vehicle using a conditional variational autoencoder},
  author={Eilermann, Sebastian and Petroll, Christoph and Hoefer, Philipp and Niggemann, Oliver},
  journal={Computer Science Research Notes},
  year={2024}
}

@article{emami2023lxgb,
  title={LXGB: a machine learning algorithm for estimating the discharge coefficient of pseudo-cosine labyrinth weir},
  author={Emami, Somayeh and Emami, Hojjat and Parsa, Javad},
  journal={Scientific Reports},
  volume={13},
  number={1},
  pages={12304},
  year={2023},
  publisher={Nature Publishing Group UK London}
}

@inproceedings{wu2023fast,
  title={Fast point cloud generation with straight flows},
  author={Wu, Lemeng and Wang, Dilin and Gong, Chengyue and Liu, Xingchao and Xiong, Yunyang and Ranjan, Rakesh and Krishnamoorthi, Raghuraman and Chandra, Vikas and Liu, Qiang},
  booktitle={Proceedings of the IEEE/CVF conference on computer vision and pattern recognition},
  pages={9445--9454},
  year={2023}
}

@article{Oertel.2015,
  title={Numerische Str{\"o}mungssimulationen von Flie{\ss}gew{\"a}ssern: Praxisanwendungen und zuk{\"u}nftige Entwicklungen},
  author={Oertel, Mario and Bung, Daniel},
  journal={Korrespondenz Wasserwirtschaft: KW},
  volume={8},
  number={H. 3},
  pages={177--182},
  year={2015},
  publisher={Gesellschaft zur F{\"o}rderung der Abwassertechnik}
}

@inproceedings{luo2021diffusion,
  title={Diffusion probabilistic models for 3d point cloud generation},
  author={Luo, Shitong and Hu, Wei},
  booktitle={Proceedings of the IEEE/CVF conference on computer vision and pattern recognition},
  pages={2837--2845},
  year={2021}
}

@article{pradas2024evaluation,
  title={Evaluation of Different Large Language Model Agent Frameworks for Design Engineering Tasks},
  author={Pradas Gomez, Alejandro and Panarotto, Massimo and Isaksson, Ola and others},
  journal={DS 130: Proceedings of NordDesign 2024, Reykjavik, Iceland, 12th-14th August 2024},
  pages={693--702},
  year={2024}
}

@inproceedings{yang2019pointflow,
  title={Pointflow: 3d point cloud generation with continuous normalizing flows},
  author={Yang, Guandao and Huang, Xun and Hao, Zekun and Liu, Ming-Yu and Belongie, Serge and Hariharan, Bharath},
  booktitle={Proceedings of the IEEE/CVF international conference on computer vision},
  pages={4541--4550},
  year={2019}
}

@inproceedings{heesch2025evaluating,
  title={Evaluating large language models for real-world engineering tasks},
  author={Heesch, Ren{\'e} and Eilermann, Sebastian and Windmann, Alexander and Diedrich, Alexander and Niggemann, Oliver},
  booktitle={Australasian Joint Conference on Artificial Intelligence},
  pages={54--66},
  year={2025},
  organization={Springer}
}

@article{vahdat2022lion,
  title={Lion: Latent point diffusion models for 3d shape generation},
  author={Vahdat, Arash and Williams, Francis and Gojcic, Zan and Litany, Or and Fidler, Sanja and Kreis, Karsten and others},
  journal={Advances in Neural Information Processing Systems},
  volume={35},
  pages={10021--10039},
  year={2022}
}

@article{eilermann2025continuous,
  title={A Continuous-Time Consistency Model for 3D Point Cloud Generation},
  author={Eilermann, Sebastian and Heesch, Ren{\'e} and Niggemann, Oliver},
  journal={arXiv preprint arXiv:2509.01492},
  year={2025}
}

@article{heidarnejad2025machine,
  title={Machine learning-based discharge coefficient estimation in trapezoidal-arched labyrinth weirs},
  author={Heidarnejad, Mohammad and Feili, Jamal and Fuladipanah, Mehdi and Rathnayake, Upaka},
  journal={Asian Journal of Water, Environment and Pollution},
  volume={22},
  number={6},
  pages={73--88},
  year={2025},
  publisher={AccScience Publishing}
}

@article{tian2024enhancing,
  title={Enhancing discharge prediction over Type-A piano key weirs: An innovative machine learning approach},
  author={Tian, Weiming and Isleem, Haytham F and Hamed, Abdelrahman Kamal and Elshaarawy, Mohamed Kamel},
  journal={Flow Measurement and Instrumentation},
  volume={100},
  pages={102732},
  year={2024},
  publisher={Elsevier}
}

@article{pedregosa2011scikit,
  title={Scikit-learn: Machine learning in Python},
  author={Pedregosa, Fabian and Varoquaux, Ga{\"e}l and Gramfort, Alexandre and Michel, Vincent and Thirion, Bertrand and Grisel, Olivier and Blondel, Mathieu and Prettenhofer, Peter and Weiss, Ron and Dubourg, Vincent and others},
  journal={Journal of machine learning research},
  volume={12},
  number={Oct},
  pages={2825--2830},
  year={2011}
}

@article{chen2016xgboost,
  title={XGBoost: A Scalable Tree Boosting System},
  author={Chen, Tianqi},
  journal={Cornell University},
  year={2016}
}

@article{ke2017lightgbm,
  title={Lightgbm: A highly efficient gradient boosting decision tree},
  author={Ke, Guolin and Meng, Qi and Finley, Thomas and Wang, Taifeng and Chen, Wei and Ma, Weidong and Ye, Qiwei and Liu, Tie-Yan},
  journal={Advances in neural information processing systems},
  volume={30},
  year={2017}
}

@article{friedman2001greedy,
  title={Greedy function approximation: a gradient boosting machine},
  author={Friedman, Jerome H},
  journal={Annals of statistics},
  pages={1189--1232},
  year={2001},
  publisher={JSTOR}
}

@article{breiman2001random,
  title={Random forests},
  author={Breiman, Leo},
  journal={Machine learning},
  volume={45},
  number={1},
  pages={5--32},
  year={2001},
  publisher={Springer}
}

@article{lundberg2017unified,
  title={A unified approach to interpreting model predictions},
  author={Lundberg, Scott M and Lee, Su-In},
  journal={Advances in neural information processing systems},
  volume={30},
  year={2017}
}

@incollection{shapley:book1952,
  title = {A Value for n-Person Games},
  author = {Shapley, Lloyd S},
  booktitle = {Contributions to the Theory of Games II},
  editor = {Kuhn, Harold W. and Tucker, Albert W.},
  pages = {307--317},
  year = {1953},
  publisher = {Princeton University Press},
  address = {Princeton}
}

@article{gharehbaghi2023comparison,
  title={A comparison of artificial intelligence approaches in predicting discharge coefficient of streamlined weirs},
  author={Gharehbaghi, Amin and Ghasemlounia, Redvan and Afaridegan, Ehsan and Haghiabi, AmirHamzeh and Mandala, Vishwanadham and Azamathulla, Hazi Mohammad and Parsaie, Abbas},
  journal={Journal of Hydroinformatics},
  volume={25},
  number={4},
  pages={1513--1530},
  year={2023},
  publisher={IWA Publishing}
}

@article{belaabed2025optimized,
  title={Optimized machine learning models for accurate prediction of the discharge coefficient in hydraulic weirs},
  author={Belaabed, Faris and Arabet, Leila and Goudjil, Kamel and Ouamane, Ahmed},
  journal={Journal of Hydrology and Hydromechanics},
  volume={73},
  number={3},
  pages={295--309},
  year={2025},
  publisher={De Gruyter Poland}
}

@article{asgharzadeh2025machine,
  title={Machine learning-based estimation of discharge coefficient for semicircular labyrinth weirs},
  author={Asgharzadeh-Bonab, Akbar and Bijanvand, Sajad and Parsaie, Abbas and Afaridegan, Ehsan},
  journal={Scientific Reports},
  volume={15},
  number={1},
  pages={33002},
  year={2025},
  publisher={Nature Publishing Group UK London}
}

@article{ef76b040-2f28-37ba-b0c4-02ed99573416,
 ISSN = {00401706},
 URL = {http://www.jstor.org/stable/1268522},
 author = {McKay, Michael D and Beckman, Richard J and Conover, William J},
 journal = {Technometrics},
 number = {2},
 pages = {239--245},
 publisher = {[Taylor & Francis, Ltd., American Statistical Association, American Society for Quality]},
 title = {A Comparison of Three Methods for Selecting Values of Input Variables in the Analysis of Output from a Computer Code},
 urldate = {2026-02-09},
 volume = {21},
 year = {1979}
}

@incollection{benesty2009pearson,
  title={Pearson correlation coefficient},
  author={Benesty, Jacob and Chen, Jingdong and Huang, Yiteng and Cohen, Israel},
  booktitle={Noise reduction in speech processing},
  pages={1--4},
  year={2009},
  publisher={Springer}
}

@article{papageorgiou2022correlation,
  title={On correlation coefficients and their interpretation},
  author={Papageorgiou, Spyridon N},
  journal={Journal of orthodontics},
  volume={49},
  number={3},
  pages={359--361},
  year={2022},
  publisher={Sage Publications Sage UK: London, England}
}

@article{mohammadi2022efficiency,
  title={Efficiency of uncertainty propagation methods for moment estimation of uncertain model outputs},
  author={Mohammadi, Samira and Cremaschi, Selen},
  journal={Computers \& Chemical Engineering},
  volume={166},
  pages={107954},
  year={2022},
  publisher={Elsevier}
}

@inproceedings{deng2009imagenet,
  title={Imagenet: A large-scale hierarchical image database},
  author={Deng, Jia and Dong, Wei and Socher, Richard and Li, Li-Jia and Li, Kai and Fei-Fei, Li},
  booktitle={2009 IEEE conference on computer vision and pattern recognition},
  pages={248--255},
  year={2009},
  organization={Ieee}
}

@inproceedings{wu20153d,
  title={3d shapenets: A deep representation for volumetric shapes},
  author={Wu, Zhirong and Song, Shuran and Khosla, Aditya and Yu, Fisher and Zhang, Linguang and Tang, Xiaoou and Xiao, Jianxiong},
  booktitle={Proceedings of the IEEE conference on computer vision and pattern recognition},
  pages={1912--1920},
  year={2015}
}

@article{zhou2016thingi10k,
  title={Thingi10k: A dataset of 10,000 3d-printing models},
  author={Zhou, Qingnan and Jacobson, Alec},
  journal={arXiv preprint arXiv:1605.04797},
  year={2016}
}

@article{torralba200880,
  title={80 million tiny images: A large data set for nonparametric object and scene recognition},
  author={Torralba, Antonio and Fergus, Rob and Freeman, William T},
  journal={IEEE transactions on pattern analysis and machine intelligence},
  volume={30},
  number={11},
  pages={1958--1970},
  year={2008},
  publisher={IEEE}
}

@inproceedings{song2023surrogate,
  title={Surrogate modeling of car drag coefficient with depth and normal renderings},
  author={Song, Binyang and Yuan, Chenyang and Permenter, Frank and Arechiga, Nikos and Ahmed, Faez},
  booktitle={International Design Engineering Technical Conferences and Computers and Information in Engineering Conference},
  volume={87301},
  pages={V03AT03A029},
  year={2023},
  organization={American Society of Mechanical Engineers}
}

@article{kashefi2022physics,
  title={Physics-informed PointNet: A deep learning solver for steady-state incompressible flows and thermal fields on multiple sets of irregular geometries},
  author={Kashefi, Ali and Mukerji, Tapan},
  journal={Journal of Computational Physics},
  volume={468},
  pages={111510},
  year={2022},
  publisher={Elsevier}
}

@article{remelli2020meshsdf,
  title={Meshsdf: Differentiable iso-surface extraction},
  author={Remelli, Edoardo and Lukoianov, Artem and Richter, Stephan and Guillard, Benoit and Bagautdinov, Timur and Baque, Pierre and Fua, Pascal},
  journal={Advances in Neural Information Processing Systems},
  volume={33},
  pages={22468--22478},
  year={2020}
}

\end{document}